\documentclass[manuscript,screen]{acmart}
\AtBeginDocument{%
  }

\setcopyright{acmlicensed}
\copyrightyear{2026}
\acmYear{2026}

\usepackage{siunitx}
\usepackage{longtable}
\usepackage{comment}
\usepackage{makecell}
\usepackage{multirow}
\usepackage[normalem]{ulem}
\useunder{\uline}{\ul}{}
\begin{document}

\title{Diffusion Models in Medical Image Inpainting: Challenges, Solution Taxonomy, and Future Directions}


\author{Arthur Dantas Mangussi}
\orcid{0000-0003-2086-532X}
\authornotemark[1]
\email{arthuradm@ita.br}
\affiliation{%
  \institution{Computer Science Division, Aeronautics Institute of Technology and Science and Technology Institute, Federal University of São Paulo}
  \city{São José dos Campos}
  \country{Brazil}
}

\author{Joana Cristo Santos}
\authornote{Both authors contributed equally to this research.}
\email{jcsantos@ciencias.ulisboa.pt}
\orcid{}
\affiliation{%
  \institution{LASIGE, Faculdade de Ciências, Universidade de Lisboa}
  \city{Lisboa}
  \country{Portugal}
}

\author{Ricardo Cardoso Pereira}
\affiliation{%
  \institution{University of Coimbra, CISUC/LASI – Centre for Informatics and Systems of the University of Coimbra, Department of Informatics Engineering}
  \city{Coimbra}
  \country{Portugal}}
\email{rdpereira@dei.uc.pt}

\author{Ana Carolina Lorena}
\affiliation{
  \institution{Computer Science Division, Aeronautics Institute of Technology and Science and Technology Institute, Federal University of São Paulo}
  \city{São José dos Campos}
  \country{Brazil}}
\email{aclorena@ita.br}

\author{Mário A. T. Figueiredo}
\affiliation{%
  \institution{Instituto Superior Técnico, Universidade de Lisboa, Instituto de Telecomunicações}
  \city{Lisboa}
  \country{Portugal}}
\email{mario.figueiredo@tecnico.ulisboa.pt}

\author{Pedro Henriques Abreu}
\affiliation{%
  \institution{University of Coimbra, CISUC/LASI – Centre for Informatics and Systems of the University of Coimbra, Department of Informatics Engineering}
  \city{Coimbra}
  \country{Portugal}}
\email{pha@dei.uc.pt}

\renewcommand{\shortauthors}{Mangussi et al.}

\begin{abstract}
  Image inpainting aims to reconstruct missing or corrupted regions of an image while preserving as much as possible, visual and semantic consistency. In medical imaging, this task is particularly important because artifacts, missing information, and pathological alterations can compromise diagnostic reliability and downstream clinical applications. Recently, diffusion models have emerged as state-of-the-art generative approaches for medical image inpainting due to their ability to generate anatomically consistent reconstructions. This survey presents a systematic review of diffusion-based methods for medical image inpainting, covering the main architectures, applications, datasets, and evaluation strategies reported across 60 studies. In addition, we propose a taxonomy for diffusion-based approaches. The analysis reveals a rapid growth of research interest in diffusion-based medical image inpainting, with denoising diffusion probabilistic models and latent diffusion models emerging as the dominant architectures. The reviewed studies mainly focus on artifact removal, data augmentation, pseudo-healthy tissue reconstruction, and anomaly detection, particularly in magnetic resonance imaging and computed tomography imaging. Overall, diffusion models demonstrate strong performance in producing anatomically plausible reconstructions and aiding downstream clinical tasks. However, the review also highlights important challenges, including the lack of standardized benchmarks, limited dataset diversity, and restricted validation procedures across diverse clinical applications and imaging scenarios.
\end{abstract}




\maketitle

\section{Introduction} \label{sec:intro}

Medical imaging has become an indispensable tool for diagnosis, monitoring, and treatment planning of a wide range of diseases, from neurological disorders detected through magnetic resonance imaging (MRI) to cardiovascular conditions assessed via computed tomography (CT) ~\cite{Joana2024,Vavekanand2026}. Despite its crucial role in modern healthcare, medical imaging remains subject to several challenges that arise from factors such as patient motion, equipment limitations, safety protocols, and time constraints during scanning~\cite{Vavekanand2026}. These factors often result in incomplete or missing data, a common issue in real-world clinical settings where ideal acquisition conditions are rarely achievable.
In this context, missing data refers to the absence of information in a particular area of an image, or even the complete loss of an image, potentially compromising its reliability and clinical usefulness~\cite{Santos2025}. This problem is particularly concerning because missing information may obscure important anatomical details, increasing the risk of misdiagnosis or treatment delays.

To address these situations, one possibility is the use of image inpainting techniques that are capable of reconstructing missing or corrupted regions of an image while preserving both visual and semantic consistency with the surrounding content \cite{Santos2025}. As a fundamental task in image processing and computer vision, image inpainting has found applications in areas such as image restoration, object removal, and video editing \cite{quan2024deep}. In healthcare, however, the challenge extends beyond visual plausibility, requiring the preservation of clinically relevant and diagnostically accurate information \cite{Santos2025}. 

Driven by recent advances in deep learning techniques, image inpainting has consequently become an active and rapidly evolving research area~\cite{Elharrouss2024}. Generative architectures such as generative adversarial networks (GAN), vision transformers (ViT), and diffusion models (DM) have been increasingly used in tasks including image reconstruction, segmentation, and inpainting~\cite{Wang2025}. Among these approaches, diffusion models have demonstrated particularly strong performance in inpainting and related generative tasks~\cite{Wang2025}. 

Recent studies show that diffusion models have been widely adopted for image generation and editing tasks, including image-to-image translation and inpainting~\cite{Croitoru2023}. Despite the rapid growth of this field, relatively few surveys have specifically focused on diffusion-based medical image inpainting. 
Existing surveys mainly focus on broad applications of diffusion models in medical imaging \cite{kazerouni2023diffusion, azad2026systematic} or on medical image synthesis \cite{khosravi2025exploring}, leaving medical image inpainting relatively unexplored. In contrast, this survey is specifically dedicated to diffusion-based medical image inpainting, combining a systematic analysis of the literature with experimental evaluations of state-of-the-art methods. By consolidating recent advances scattered across numerous publications, this work identifies key methodologies, trends, challenges, and future research opportunities, aiming to serve as a valuable resource for researchers and practitioners in the field.

In summary, the main contributions of this work are:

\begin{itemize}
\item A systematic review of diffusion-based medical image inpainting methods and recent research trends, identifying the most widely adopted architectures, applications, and imaging modalities;

\item A taxonomy of diffusion-based approaches for medical imaging applications, providing a structured organization of existing methodologies and facilitating future research in the field;

\item A comparative analysis of datasets, evaluation metrics, and experimental setups, aiming to identify common evaluation practices and contribute toward a more standardized assessment methodology;

\item An experimental case study evaluating representative state-of-the-art methods, enabling a more detailed comparison of diffusion-based inpainting methodologies.
\end{itemize}

The remainder of this paper is organized as follows. Section~\ref{sec:methodology} describes the methodology adopted for conducting the systematic review, while Section~\ref{sec:background_knowledge} provides the overview and taxonomy of diffusion models. Section~\ref{sec:results} presents the main findings of the literature review and the key patterns identified. Subsequently, Section~\ref{sec:discussion} provides an in-depth discussion of diffusion models applied to medical image inpainting. Section~\ref{sec:case_study} introduces a case study illustrating the practical application of these methods, and finally, Section~\ref{sec:conclusions} concludes the paper.

\section{Methodology} \label{sec:methodology}

The primary objective is to analyze recent advances in the application of diffusion models to medical image inpainting. To achieve this, a systematic literature search was conducted across multiple databases such as Scopus, Web of Science, and IEEE Xplore databases using combinations of the keywords \textit{image inpainting, diffusion models, image completion, medical images, clinical, healthcare,} and \textit{radiology}. The search covered publications published between 2023 and 2026, since DDPMs were first officially presented in 2020 \cite{Ho2020}. Initially, a total of 841 works were retrieved. 

\begin{figure}[ht]
    \centering
    \includegraphics[width=0.8\linewidth]{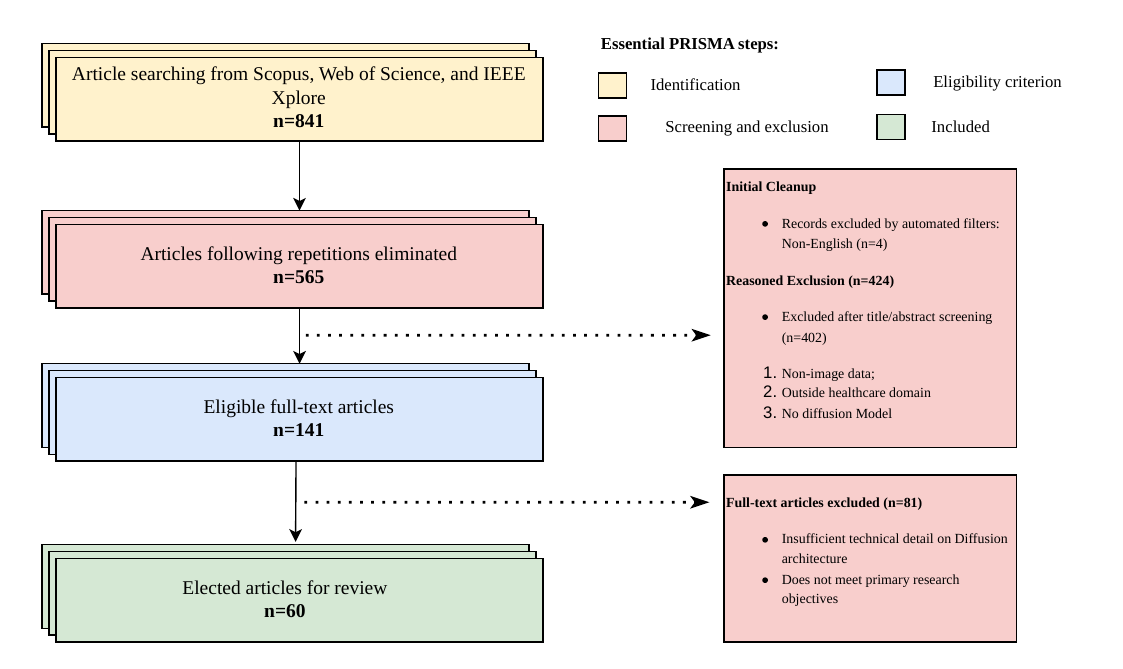}
    \caption{The employed PRISMA method for the article searching procedure, where we demonstrate the evidence for the article’s inclusion and exclusion criteria.}
    \label{fig:prisma}
    \Description[PRISMA method]{The employed PRISMA method for article search, where we demonstrate the evidence for the article’s inclusion and exclusion criteria.}
\end{figure}

This review follows the well-established Preferred Reporting Items for Systematic Reviews and Meta-Analyses (PRISMA) framework~\cite{Hasan2021}, as illustrated in Figure~\ref{fig:prisma}. PRISMA provides an evidence-based set of guidelines designed to improve the transparency and completeness of systematic reviews and meta-analyses, and is widely used to structure literature selection and reporting~\cite{Hasan2021}.

During the identification stage, duplicate records were removed (n = 276), resulting in 565 works eligible for the screening phase. Subsequently, title and abstract screening were performed to assess the relevance of the retrieved studies. Works were excluded if they were not written in English, did not employ diffusion-based architectures, were unrelated to image data, or did not address image inpainting tasks. After this initial screening process, 141 studies remained eligible for full-text assessment.

During the eligibility decision stage, the full texts of the remaining works were carefully evaluated to determine their alignment with the central scope of this survey, namely the application of diffusion models to medical image inpainting. Works were excluded if they focused on non-medical imaging applications, employed diffusion models for tasks unrelated to inpainting, or addressed adjacent but distinct problems, such as general image generation~\cite{Hung2023} or missing modality synthesis~\cite{Kalantar2023, Kebaili2025}.

Following the complete PRISMA workflow, a total of 60 works satisfied all inclusion criteria and were ultimately selected for this review. 

\section{Diffusion Models: Overview and Proposed Taxonomy}
\label{sec:background_knowledge}

 In recent years, generative artificial intelligence (GenAI) has experienced a rapid advancement, driven by both architectural innovations and increased computational capabilities \cite{10521640}. According to Feuerriegel et al.~\cite{Feuerriegel2023}, the term ``generative AI'' refers to computational techniques capable of producing new, meaningful content, such as text, images, or audio, by learning underlying data distributions. Prominent model families include generative adversarial networks (GAN), autoencoders (AE), transformers, and diffusion models~\cite{sengar2025generative}. Table~\ref{tab:generative_models} summarizes the main characteristics of these architectures.

\begin{table}[!ht]
    \centering
    \caption{Overview of core deep generative model architectures~\cite{Banh2023}.}
    \label{tab:generative_models}
    \small
    \begin{tabular}{p{2.2cm} p{10.5cm}}
        \toprule
        \textbf{Architecture} & \textbf{Description} \\
        \midrule
        \textbf{AE} & \begin{tabular}[t]{@{}l@{}}Employs an encoder-decoder framework to map data into a latent space and reconstruct it. \\ Learning is performed are generated by optimizing a lower bound on the data likelihood \\ of a training set. \end{tabular}\\
        \addlinespace
        \textbf{GAN} & \begin{tabular}[t]{@{}l@{}}Consists of a generator and a discriminator trained in an adversarial setting. \\ The generator synthesizes data, while the discriminator distinguishes between real and \\ generated samples, driving the generation of high-quality outputs.\end{tabular} \\
        \addlinespace
        \textbf{Transformer} & \begin{tabular}[t]{@{}l@{}}Uses self-attention mechanisms to capture long-range dependencies, forming the\\ foundation of modern large-scale models for sequential and multimodal data.\end{tabular} \\
        \addlinespace
        \textbf{Diffusion} & \begin{tabular}[t]{@{}l@{}}Defines a forward process that progressively adds Gaussian noise over $T$ steps, and a \\ learned reverse process (typically parameterized by a U-Net) that is trained to\\ reconstruct the original data through iterative denoising.\end{tabular}  \\
        \bottomrule
    \end{tabular}
\end{table}

As highlighted in Table~\ref{tab:generative_models}, the U-Net architecture has become the \textit{de-facto} choice of backbone for standard diffusion models for images. First introduced in this context by Ho et al., this convolutional architecture is primarily comprised of ResNet blocks interspersed with spatial self-attention modules at lower resolutions~\cite{Ho2020}. However, recent research has demonstrated that the specific inductive bias of the U-Net is not strictly crucial to the performance of diffusion models. Consequently, emerging architectures, such as diffusion transformers (DiTs), have successfully replaced the traditional U-Net backbone with standard transformer blocks that operate on latent patches~\cite{Peebles2023}.

\begin{figure}[!htbp]
    \centering
    \includegraphics[scale=0.5]{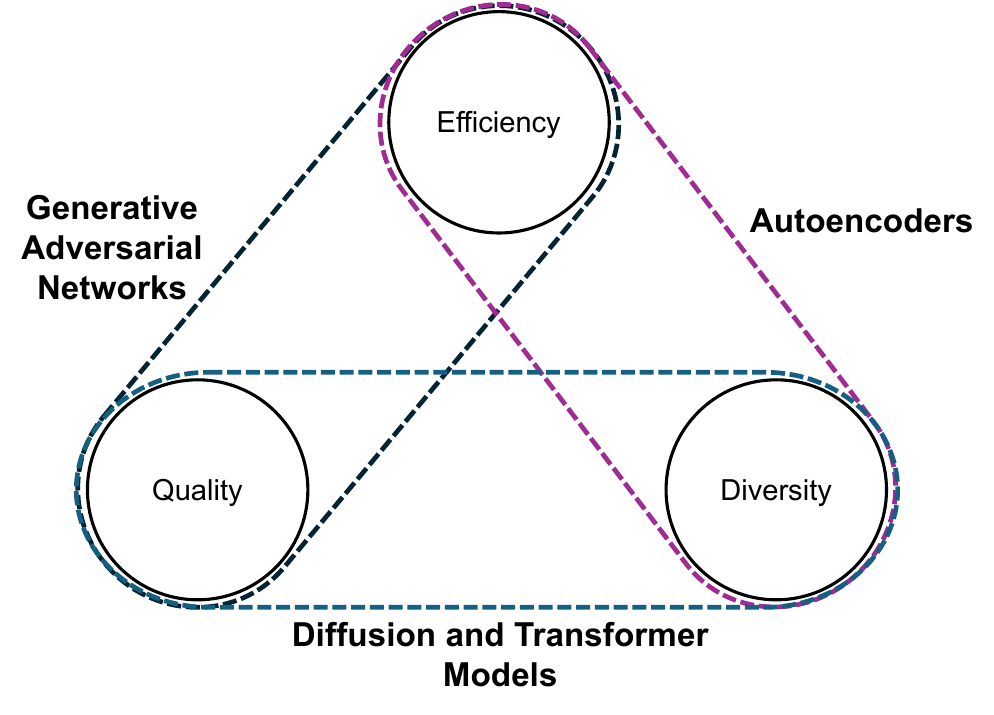}
    \caption{Trade-off between sample quality, diversity, and computational efficiency across major generative model families.}
    \label{fig:trilemma}
    \Description[]{}
\end{figure}

While GANs are capable of generating high-fidelity samples efficiently, they often suffer from limited mode coverage and are famously difficult to train ~\cite{xiao2022tackling}. Autoencoders, particularly variational autoencoders (VAEs), typically achieve better diversity but may produce lower-quality samples. Diffusion and transformer-based models have emerged as compelling alternatives, offering a favorable trade-off between sample quality and mode coverage, although often at the expense of computational efficiency. In particular, the iterative sampling process of diffusion models and the high computational complexity of transformers contribute to slower inference and increased resource demands~\cite{kazerouni2023diffusion}. Figure~\ref{fig:trilemma} illustrates this trade-off among generative model families.

Diffusion models have recently achieved state-of-the-art performance in medical image inpainting for several reasons. First, their training objective, based on denoising score matching, is stable and well-defined, avoiding adversarial training instabilities and issues such as mode collapse, commonly associated with GANs. Second, unlike traditional AE-based generative approaches, diffusion models do not include restrictive latent-space assumptions, enabling them to better capture complex and multimodal anatomical distributions better while producing sharper reconstructions. Third, compared with transformer-based architectures, diffusion models benefit from a strong inductive bias toward local spatial structures through their U-Net-based designs, while still incorporating global contextual information via attention mechanisms in hybrid architectures.

\begin{figure}[!htbp]
    \centering
    \includegraphics[scale=0.55]{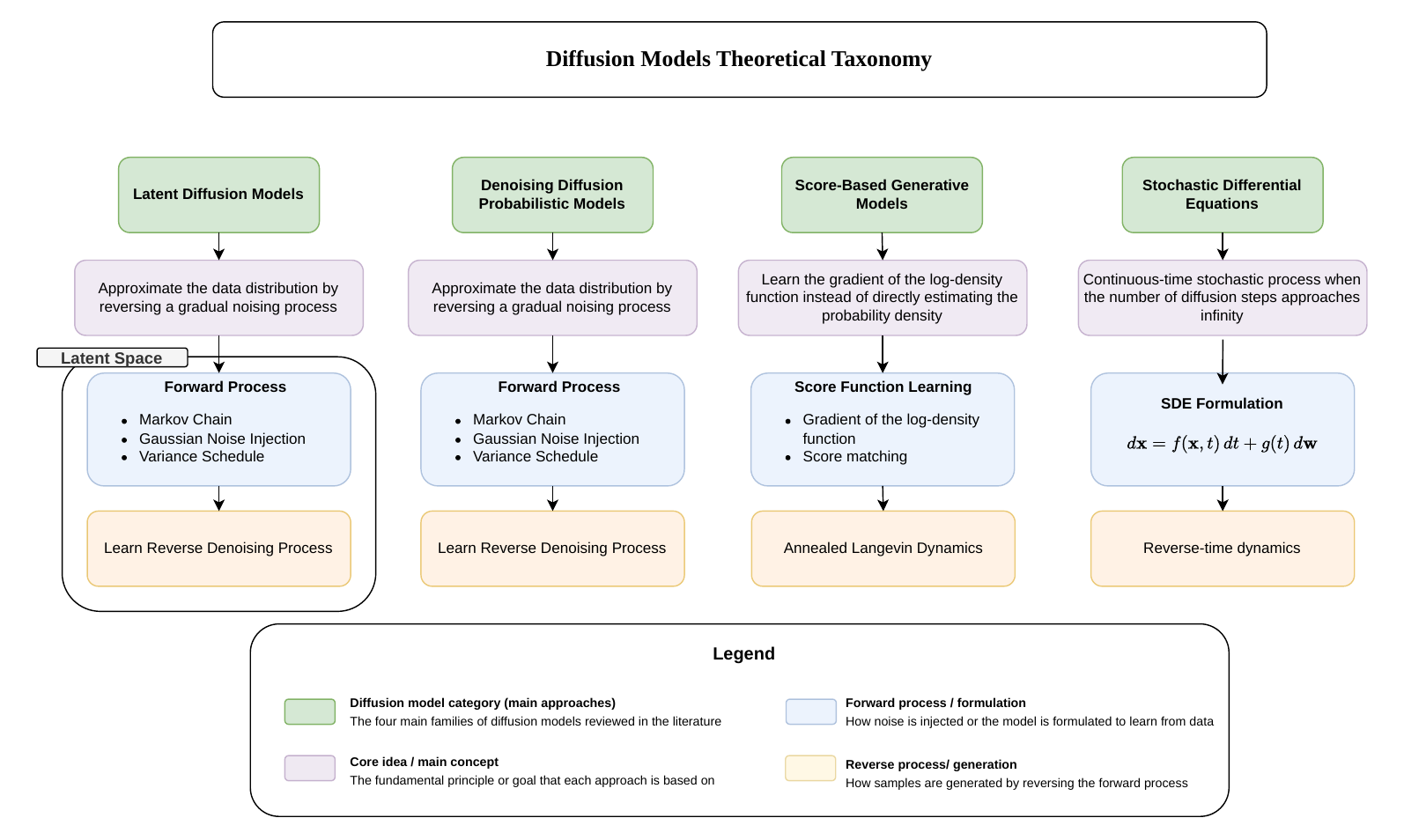}
    \caption{Proposed taxonomy of diffusion-based models.}
    \label{fig:taxonomy}
    \Description[]{}
\end{figure}

\begin{figure}[!htbp]
    \centering
    \includegraphics[scale=0.6]{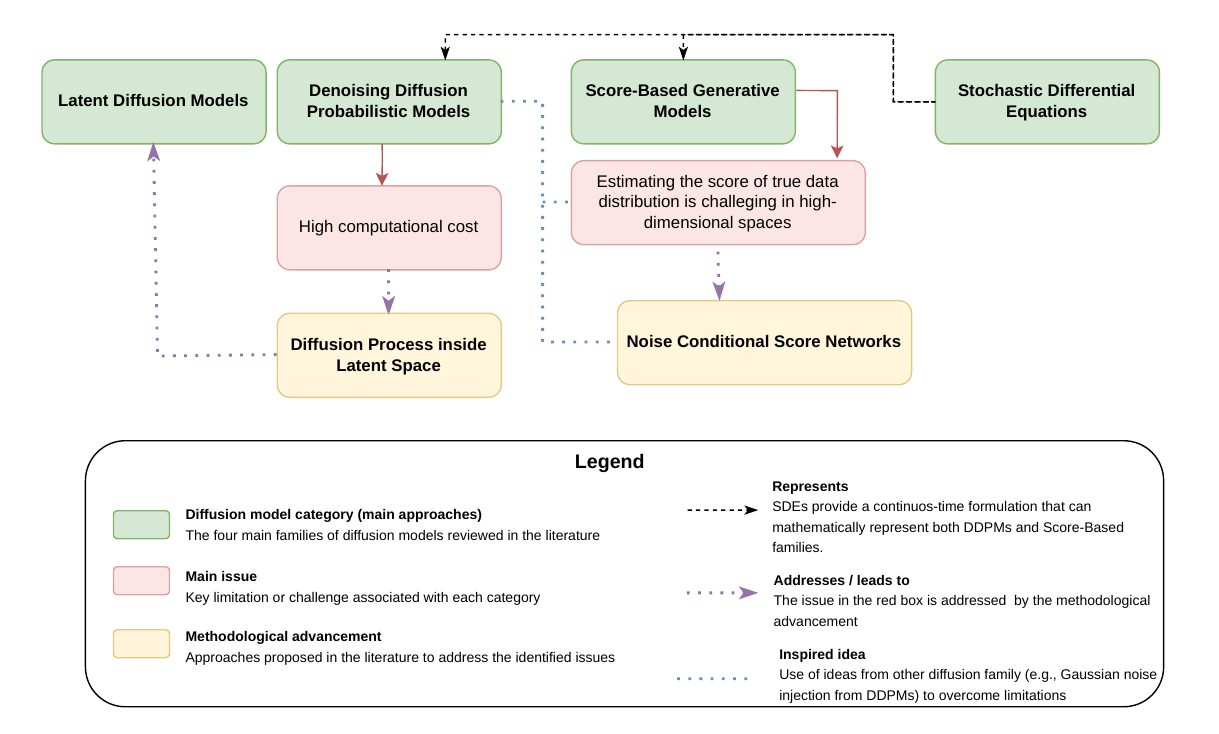}
    \caption{Conceptual workflow of diffusion model families and their methodological evolution.}
    \label{fig:taxonomy_flow}
    \Description[]{}
\end{figure}

Within the broader diffusion model literature, several formulations have been proposed, differing in their mathematical foundations, noise parameterizations, and inference procedures. The following sections review the main families of diffusion models and summarize their theoretical foundations. Figure~\ref{fig:taxonomy} presents the proposed taxonomy of diffusion-based models, highlighting the categories and their theoretical relationships, while Figure~\ref{fig:taxonomy_flow} illustrates the conceptual workflow of these approaches and how the literature addressed the main methodological limitations of each family throughout the evolution of diffusion models.

\subsection{Denoising Diffusion Probabilistic Models}

\begin{figure}[!htbp]
    \centering
    \includegraphics[scale=0.75]{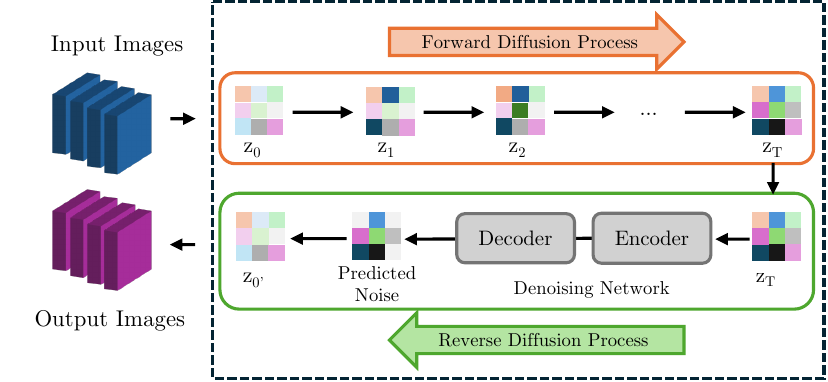}
    \caption{Denoising diffusion probabilistic model (DDPM) Architecture.}
    \label{fig:ddpm}
    \Description[]{}
\end{figure}

Denoising diffusion probabilistic models (DDPMs) are generative models that learn to approximate the data distribution by reversing a gradual noising process. The framework consists of two Markov processes: a forward (diffusion) process $q$, which progressively corrupts the data by adding Gaussian noise, and a learned reverse (denoising) process $p_\theta$, which is trained to reconstruct the data by removing noise, as illustrated in Figure \ref{fig:ddpm}.

The forward process is defined as a Markov chain that transforms an input sample $\mathbf{x}_0$ into a sequence of increasingly noisy variables $\mathbf{x}_1, \dots, \mathbf{x}_T$ over $T$ time steps,
\begin{equation}
q(\mathbf{x}_t \mid \mathbf{x}_{t-1}) = \mathcal{N}(\mathbf{x}_t; \sqrt{1 - \beta_t}\,\mathbf{x}_{t-1}, \beta_t \mathbf{I}),
\end{equation}
where $\beta_t \in (0,1)$ is a predefined variance schedule controlling the amount of noise added at each step. Defining $\alpha_t = 1 - \beta_t$ and the cumulative product $\bar{\alpha}_t = \prod_{s=1}^{t} \alpha_s$, the forward process admits a closed-form expression:
\begin{equation}
q(\mathbf{x}_t \mid \mathbf{x}_0) = \mathcal{N}(\mathbf{x}_t; \sqrt{\bar{\alpha}_t}\,\mathbf{x}_0, (1 - \bar{\alpha}_t)\mathbf{I}).
\end{equation}

\noindent
This formulation allows direct sampling from any timestep $t$ using the reparameterization:
\begin{equation}
\mathbf{x}_t = \sqrt{\bar{\alpha}_t}\,\mathbf{x}_0 + \sqrt{1 - \bar{\alpha}_t}\,\boldsymbol{\epsilon}, \quad \boldsymbol{\epsilon} \sim \mathcal{N}(\mathbf{0}, \mathbf{I}).
\end{equation}
As $t$ increases, $\bar{\alpha}_t \rightarrow 0$, and $\mathbf{x}_T$ approaches white Gaussian noise.

The reverse process aims to invert this noising procedure. Since the true reverse distribution $q(\mathbf{x}_{t-1} \mid \mathbf{x}_t, \mathbf{x}_0)$ is intractable, DDPMs learn a parameterized approximation,
\begin{equation}
p_\theta(\mathbf{x}_{t-1} \mid \mathbf{x}_t) = \mathcal{N}(\mathbf{x}_{t-1}; \boldsymbol{\mu}_\theta(\mathbf{x}_t, t), \sigma_t^2 \mathbf{I}),
\end{equation}
where $\boldsymbol{\mu}_\theta$ is predicted by a neural network (with parameters $\theta$) and $\sigma_t^2$ is either fixed or learned. Following Ho et al. ~\cite{Ho2020}, the mean is commonly parameterized in terms of noise prediction,
\begin{equation}
\boldsymbol{\mu}_\theta(\mathbf{x}_t, t) =
\frac{1}{\sqrt{\alpha_t}} \left(
\mathbf{x}_t - \frac{1 - \alpha_t}{\sqrt{1 - \bar{\alpha}_t}} \boldsymbol{\epsilon}_\theta(\mathbf{x}_t, t)
\right),
\end{equation}
where $\boldsymbol{\epsilon}_\theta(\mathbf{x}_t, t)$ denotes the model's estimate of the noise added at step $t$.

The model is trained by minimizing a simplified objective derived from the variational evidence lower bound (ELB),
\begin{equation}
\mathcal{L}_{\text{simple}} =
\mathbb{E}_{t, \mathbf{x}_0, \boldsymbol{\epsilon}} \left[
\left\| \boldsymbol{\epsilon} - \boldsymbol{\epsilon}_\theta(\mathbf{x}_t, t) \right\|^2
\right].
\end{equation}
This objective reduces the learning problem to predicting the noise component, enabling stable and efficient training.

At generation time, new samples are obtained by starting from Gaussian noise $\mathbf{x}_T \sim \mathcal{N}(\mathbf{0}, \mathbf{I})$ and iteratively applying the learned reverse transitions:
\begin{equation}
\mathbf{x}_{t-1} \sim p_\theta(\mathbf{x}_{t-1} \mid \mathbf{x}_t),
\end{equation}
until a sample $\mathbf{x}_0$ is produced.

\subsection{Latent Diffusion Models}

\begin{figure}[!ht]
    \centering
    \includegraphics[scale=0.75]{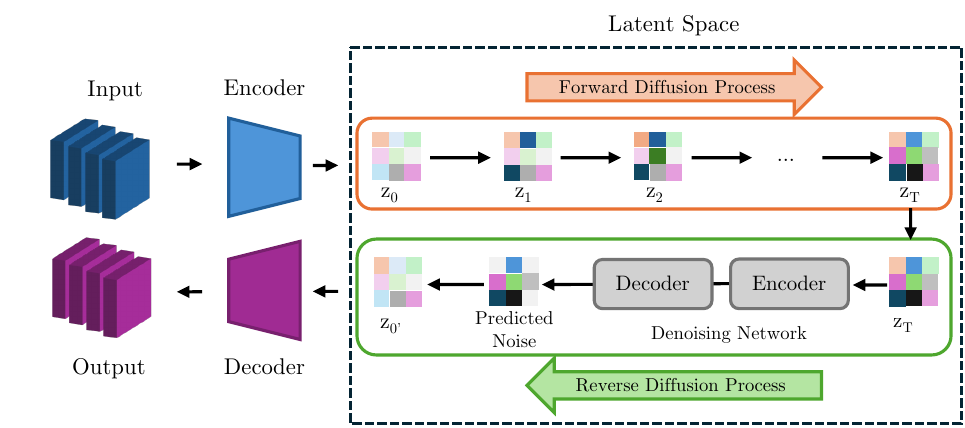}
    \caption{Architecture of the Latent Diffusion Model (LDM).}
    \label{fig:latent}
    \Description[]{}
\end{figure}

Latent diffusion models (LDMs), introduced by Rombach et al.~\cite{Rombach2022}, address the high computational cost of standard diffusion models by performing the denoising process in a lower-dimensional latent space rather than directly in pixel space (Figure~\ref{fig:latent}). This design significantly reduces computational overhead while preserving high perceptual quality.

LDMs rely on a pre-trained autoencoder for perceptual compression. An encoder $E$ maps an input image $\mathbf{x}$ to a latent representation $\mathbf{z} = E(\mathbf{x})$, while a decoder $D$ reconstructs the image as $\tilde{\mathbf{x}} = D(\mathbf{z})$.
The diffusion process is then applied in the latent space. The training objective follows the standard noise prediction formulation,
\begin{equation}
\mathcal{L}_{\text{LDM}} = \mathbb{E}_{\mathbf{z} \sim E(\mathbf{x}), \ \boldsymbol{\epsilon} \sim \mathcal{N}(0, \mathbf{I}), \ t} 
\left[
\left\| \boldsymbol{\epsilon} - \boldsymbol{\epsilon}_\theta(\mathbf{z}_t, t) \right\|_2^2
\right],
\end{equation}
\noindent
where $\mathbf{x}$ denotes the input image, $E(\mathbf{x})$ is the encoder that maps the image into the latent representation $\mathbf{z}$, and $\boldsymbol{\epsilon} \sim \mathcal{N}(0,\mathbf{I})$ represents Gaussian noise sampled from a standard normal distribution. The variable $t$ corresponds to the diffusion timestep, while $\mathbf{z}_t$ denotes the noisy latent representation obtained after adding noise at timestep $t$. The network $\boldsymbol{\epsilon}_\theta(\mathbf{z}_t,t)$ is trained to predict the injected noise, and the objective minimizes the mean squared error between the true and predicted noise.

Rombach et al.~\cite{Rombach2022} extend LDMs to the conditional setting, enabling greater flexibility by incorporating external information into the generation process. This is achieved by augmenting the U-Net backbone with cross-attention mechanisms, which allow the model to effectively integrate signals from different modalities.

\begin{figure}[!ht]
    \centering
    \includegraphics[scale=0.75]{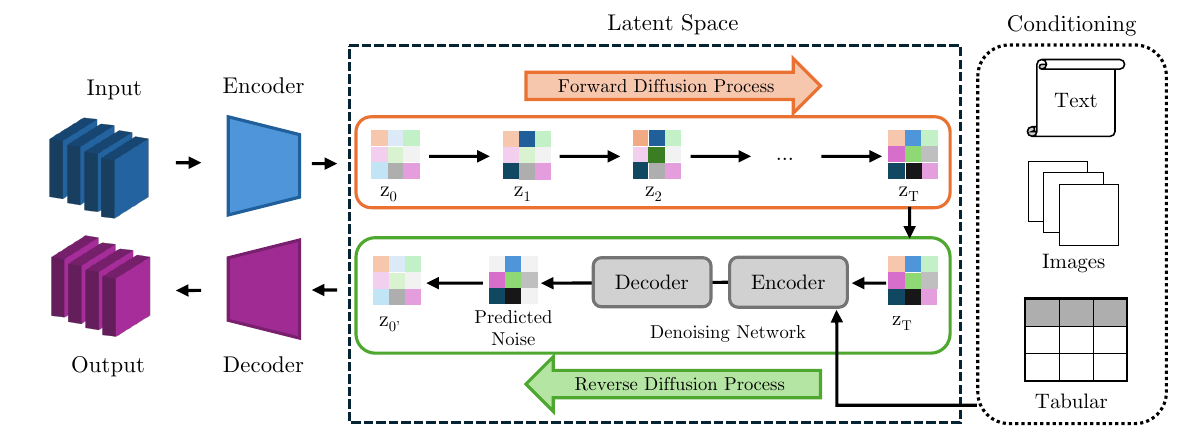}
    \caption{Architecture of the Conditional Latent Diffusion Model.}
    \label{fig:conditioning}
    \Description[]{}
\end{figure}

To incorporate conditioning information (e.g., text prompts), a domain-specific encoder $\tau_\theta$ maps the input $\mathbf{y}$ into an intermediate representation $\tau_\theta(\mathbf{y}) \in \mathbb{R}^{M \times d_\tau}$. This representation is injected into the U-Net through cross-attention layers (Figure~\ref{fig:conditioning}).
Specifically, at layer $i$, attention is computed as:
\begin{equation}
\mathrm{Attention}(\mathbf{Q}, \mathbf{K}, \mathbf{V}) = \mathrm{softmax}\left( \frac{\mathbf{Q}\mathbf{K}^\top}{\sqrt{d}} \right)\mathbf{V},
\end{equation}
where
\begin{equation}
\mathbf{Q} = \mathbf{W}_Q^{(i)} \, \phi_i(\mathbf{z}_t), 
\mathbf{K} = \mathbf{W}_K^{(i)} \, \tau_\theta(\mathbf{y}), 
\mathbf{V} = \mathbf{W}_V^{(i)} \, \tau_\theta(\mathbf{y}).
\end{equation}

\noindent
Here, $\phi_i(\mathbf{z}_t) \in \mathbb{R}^{N \times d_\epsilon^i}$ denotes the flattened intermediate feature map of the U-Net at layer $i$, and $\mathbf{W}_Q^{(i)} \in \mathbb{R}^{d \times d_\epsilon^i}$, $\mathbf{W}_K^{(i)}, \mathbf{W}_V^{(i)} \in \mathbb{R}^{d \times d_\tau}$ are learnable projection matrices.
The conditional training objective is then defined as:
\begin{equation}
\mathcal{L}_{\text{LDM}} = \mathbb{E}_{\mathbf{z} \sim E(\mathbf{x}), \ \boldsymbol{\epsilon} \sim \mathcal{N}(0, \mathbf{I}), \ t} 
\left[
\left\| \boldsymbol{\epsilon} - \boldsymbol{\epsilon}_\theta(\mathbf{z}_t, t, \tau_\theta(\mathbf{y})) \right\|_2^2
\right].
\end{equation}

\subsection{Score-Based Generative Models}

The objective of explicit generative models is to learn the underlying probability distribution of the data and subsequently generate new samples through sampling procedures~\cite{Chen2025}. Unlike conventional likelihood-based approaches, Score-Based Generative Models (SGMs) learn the gradient of the log-density function, known as the score function, instead of directly estimating the probability density itself~\cite{SongErmon2019}.

The score function is defined as:
\begin{equation}
\label{eq:log_grad}
\nabla_{\mathbf{x}} \log p(\mathbf{x})
=
\frac{\partial \log p(\mathbf{x})}{\partial \mathbf{x}}.
\end{equation}

Given a data distribution $p(\mathbf{x})$, the score function indicates the direction of maximum increase in data likelihood. SGMs approximate this quantity through a neural network $s_\theta(\mathbf{x})$:
\begin{equation}
s_\theta(\mathbf{x})
\approx
\nabla_{\mathbf{x}} \log p(\mathbf{x}).
\end{equation}

The score network can also be parameterized through an energy-based formulation:
\begin{equation}
p_\theta(\mathbf{x})
=
\frac{e^{-f_\theta(\mathbf{x})}}{Z_\theta},
\end{equation}
where $f_\theta(\mathbf{x})$ denotes the energy function and
\begin{equation}
Z_\theta
=
\int e^{-f_\theta(\mathbf{x})} d\mathbf{x},
\end{equation}
is the partition function.

Under this formulation, the score function becomes:
\begin{equation}
s_\theta(\mathbf{x})
=
\nabla_{\mathbf{x}} \log p_\theta(\mathbf{x})
=
-\nabla_{\mathbf{x}} f_\theta(\mathbf{x}),
\end{equation}
since $\nabla_{\mathbf{x}} \log Z_\theta = 0$, as $Z_\theta$ is independent of $\mathbf{x}$. This property is particularly advantageous because it removes the need to explicitly compute the normalization constant, greatly increasing the flexibility of model design.

SGMs are commonly trained through score matching, which minimizes the Fisher divergence between the model distribution and the true data distribution. The score matching objective can be expressed as~\cite{SongErmon2019}:
\begin{equation}
\mathcal{L}_{\text{SM}}
=
\mathbb{E}_{p(\mathbf{x})}
\left[
\mathrm{Tr}
\left(
\nabla_{\mathbf{x}} s_\theta(\mathbf{x})
\right)
+
\frac{1}{2}
\left\|
s_\theta(\mathbf{x})
\right\|_2^2
\right],
\end{equation}

where $\nabla_xs_\theta(x)$ denotes the Jacobian of $s_\theta$.
However, directly estimating the score of the true data distribution is challenging in high-dimensional spaces. To address this issue, Song and Ermon~\cite{SongErmon2019} proposed Noise Conditional Score Networks (NCSNs), in which the score network is trained across multiple noise scales by perturbing data samples with Gaussian noise. This multi-scale strategy smooths the data distribution, stabilizes score estimation, and improves coverage of low-density regions.

After training, samples are generated through Langevin Dynamics, which iteratively refines noisy samples using the estimated score function:
\begin{equation}
\mathbf{x}_{t}
=
\mathbf{x}_{t-1}
+
\frac{\epsilon}{2}
s_\theta(\mathbf{x}_{t-1})
+
\sqrt{\epsilon}\mathbf{z}_{t},
\end{equation}
where $\epsilon$ is the step size and $\mathbf{z}_t \sim \mathcal{N}(0,\mathbf{I})$ denotes Gaussian noise.

Furthermore, SGMs employ Annealed Langevin Dynamics, where sampling starts from high noise levels and progressively reduces the noise magnitude during generation. This procedure enables the transformation of Gaussian noise into realistic samples and establishes an important theoretical connection between score matching methods and diffusion probabilistic models.

\subsection{Stochastic Differential Equations}

The forward noise perturbation process in SGMs and diffusion models can be formulated as a continuous-time stochastic process if the number of diffusion steps approaches infinity. Many stochastic processes can be represented through stochastic differential equations (SDEs), which are generally formulated as

\begin{equation}
\label{eq:sde}
d\mathbf{x} = f(\mathbf{x}, t)\,dt + g(t)\,d\mathbf{w},
\end{equation}
where $f(\mathbf{\cdot}, t): \mathbb{R}^d \rightarrow \mathbb{R}^d$ is the drift coefficient at time $t$, $g(t) \in \mathbb{R}$ is the diffusion coefficient, $d\mathbf{w}$ denotes infinitesimal Gaussian noise, and $\mathbf{w}$ is standard Brownian motion.

\begin{figure}[!ht]
    \centering
    \includegraphics[scale=0.75]{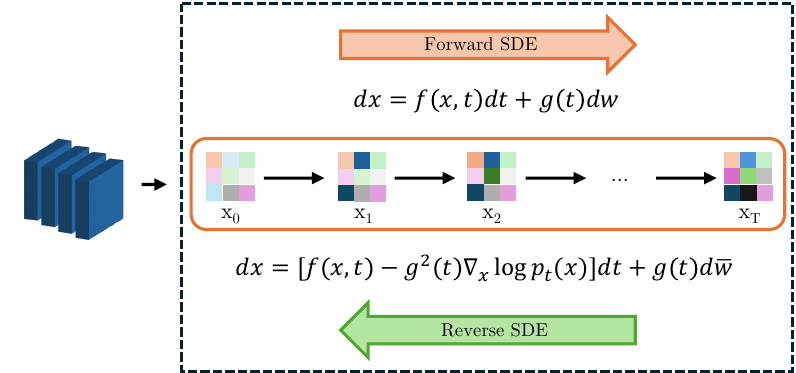}
    \caption{Architecture of the Stochastic Differential Equations Model.}
    \label{fig:sde}
    \Description[]{}
\end{figure}

Anderson~\cite{ANDERSON1982} showed that the reverse-time dynamics of an SDE can be expressed as:
\begin{equation}
d\mathbf{x} =
\left[
f(\mathbf{x}, t)
-
g^2(t)\nabla_{\mathbf{x}}\log p_t(\mathbf{x})
\right]dt
+
g(t)\,d\mathbf{w},
\end{equation}
where $dt$ is a negative infinitesimal time step in the reverse process, and $\nabla_{\mathbf{x}}\log p_t(\mathbf{x})$ is the score function (Figure \ref{fig:sde}).

In practice, the score function is approximated by a neural network $s_\theta(\mathbf{x}, t)$. The continuous-time training objective is defined through the Fisher divergence
\begin{equation}
\mathbb{E}_{t \in \mathcal{U}(0,T)}
\mathbb{E}_{p_t(\mathbf{x})}
\left[
\lambda(t)
\left\|
\nabla_{\mathbf{x}}\log p_t(\mathbf{x})
-
s_\theta(\mathbf{x}, t)
\right\|_2^2
\right],
\end{equation}
where $\mathcal{U}(0,T)$ denotes the uniform distribution over the interval $[0,T]$, and $\lambda(t)$ is a weighting function.

Song et al.~\cite{song2019sliced} demonstrated that both DDPMs and SGMs can be interpreted as particular instances of SDEs with different forward noise processes. DDPMs correspond to the variance preserving SDE (VP-SDE), defined as:
\begin{equation}
d\mathbf{x}
=
-\frac{1}{2}\beta(t)\mathbf{x}\,dt
+
\sqrt{\beta(t)}\,d\mathbf{w},
\end{equation}
where $\beta(t)$ is a predefined variance schedule~\cite{Chen2025}.

In contrast, SGMs correspond to the variance exploding SDE (VE-SDE):
\begin{equation}
d\mathbf{x}
=
\sqrt{\frac{d[\sigma^2(t)]}{dt}}\,d\mathbf{w},
\end{equation}
where $\sigma(t)$ controls the noise magnitude over time~\cite{Chen2025}. Therefore, both diffusion probabilistic models and score-based generative models can be unified under the continuous-time SDE framework, differing primarily in their choice of drift and diffusion processes.


\section{Results}
\label{sec:results}

This section presents the main findings, trends, and patterns identified in the literature review regarding the application of diffusion models to medical image inpainting. First, an overall analysis of the reviewed studies is provided, followed by a discussion structured according to the identified trends of diffusion-based approaches.

\subsection{Overall Analysis}

\begin{figure}[ht]
    \centering
    \includegraphics[scale=0.4]{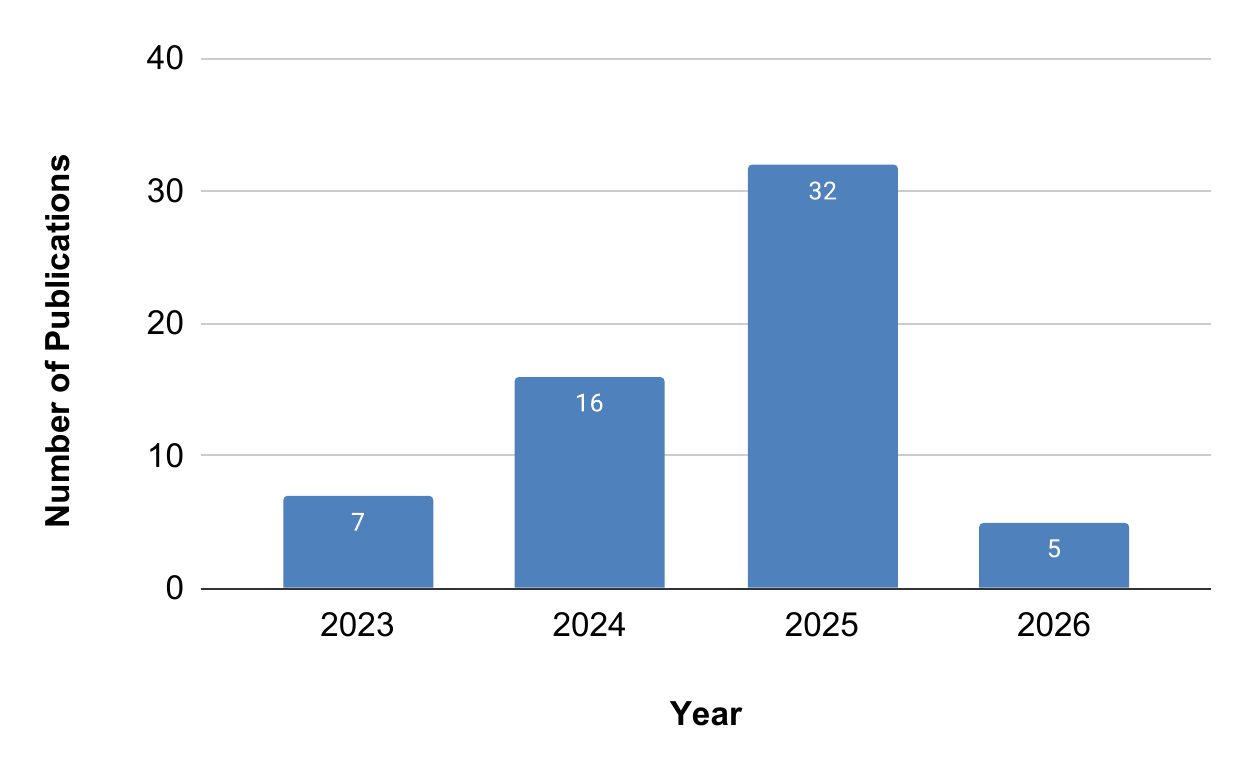}
    \caption{Number of Publications per Year on Diffusion Models for Medical Image Inpainting.}
    \label{fig:hist}
    \Description[]{}
\end{figure}

As illustrated in Figure~\ref{fig:hist}, the number of publications on diffusion-based medical image inpainting increased steadily between 2023 and 2025, highlighting the rapid growth of research interest in this area. 

\begin{figure}[!htbp]
    \centering
    \includegraphics[scale=0.4]{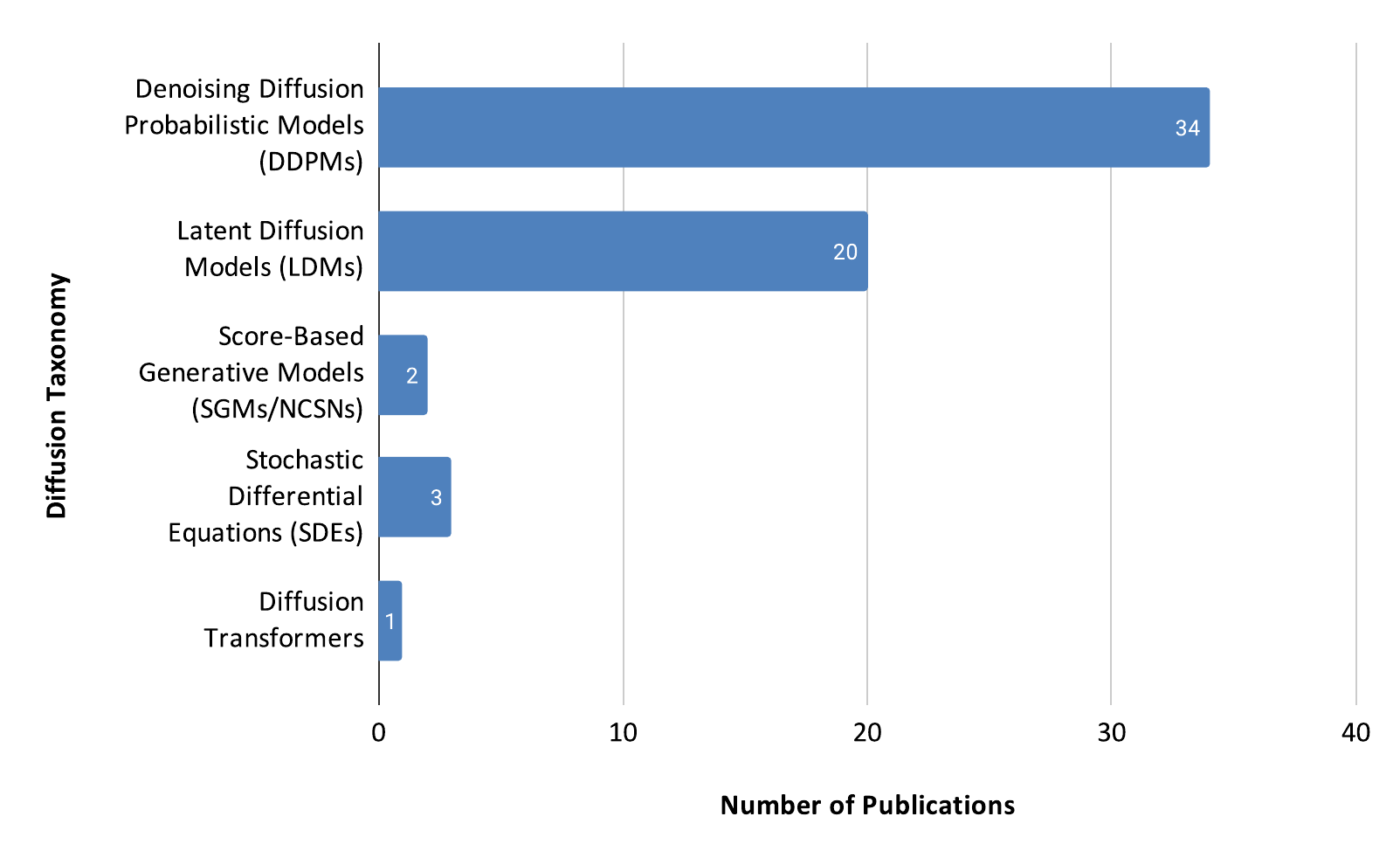}
    \caption{Proportion of Diffusion Model Families Across Reviewed Studies.}
    \label{fig:families}
    \Description[]{}
\end{figure}

The increasing adoption of diffusion-based approaches is further reflected in the distribution of model families shown in Figure~\ref{fig:families}. Among the reviewed studies, DDPMs and LDMs emerged as the dominant architectures, accounting for 56.7\% and 33.3\% of the analyzed works, respectively. 

\begin{figure}[!ht]
    \centering
    \includegraphics[scale=0.4]{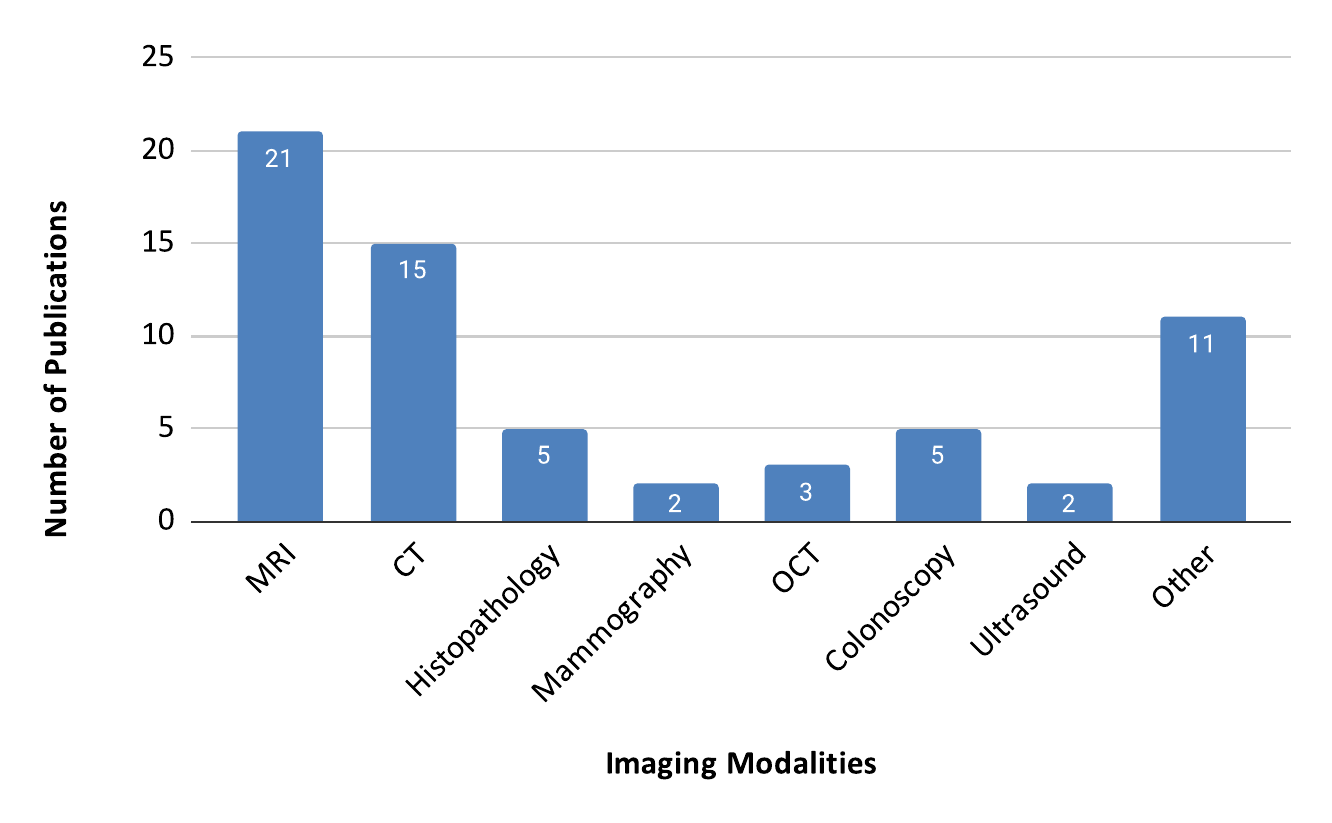}
    \caption{Distribution of Imaging Modalities Across Reviewed Studies.}
    \label{fig:modalities}
    \Description[]{}
\end{figure}

As discussed previously, diffusion models have been applied to a broad range of healthcare applications spanning multiple medical imaging modalities. Figure~\ref{fig:modalities} shows that computed tomography (CT) and magnetic resonance imaging (MRI) are the most commonly investigated modalities, while a smaller number of studies explore additional modalities such as histopathology, optical coherence tomography (OCT), and ultrasound.

With respect to image inpainting evaluation, this survey identifies several quantitative metrics commonly used to assess reconstruction quality and downstream clinical tasks such as segmentation and classification, as summarized in Table~\ref{tab:metrics}. Among the reviewed studies, the structural similarity index measure (SSIM) and peak signal-to-noise ratio (PSNR) were the most frequently reported metrics, appearing in 53.3\% of the studies, followed by the Dice similarity coefficient (DSC), reported in 28.3\% of the works. Although the remaining metrics appear less frequently, they still provide valuable complementary insights into the performance and clinical relevance of the evaluated approaches.

\begin{table}[!ht]
\centering
\footnotesize 

\caption{Image quality metrics commonly used for evaluation in medical image inpainting.
$I$ denotes the inpainted image, $\hat{I}$ the ground truth, $N$ the total number of pixels,
$\mu$ and $\sigma$ the mean and standard deviation of pixel intensities,
and $\sigma_{I\hat{I}}$ the covariance between the two images.}
\label{tab:metrics}
\begin{tabular}{p{2.5cm} p{6.5cm} p{5cm}}
\toprule
\textbf{Metric} & \textbf{Formulation} & \textbf{Best Performance} \\
\midrule

\textbf{PSNR}
\newline {\small Peak Signal-to-Noise Ratio}
&
$\mathrm{PSNR} = 10 \cdot \log_{10} \!\left(\dfrac{\mathrm{MAX}^{2}}{\mathrm{MSE}}\right)$
\newline\smallskip
{\small where $\mathrm{MAX}$ is the maximum possible pixel value and $\mathrm{MSE}$ is the Mean Squared Error between $I$ and $\hat{I}$.}
&
\textbf{Higher is better.} \newline No fixed upper bound; values $\geq 30$\,dB are generally considered high quality in medical imaging. \\

\midrule

\textbf{SSIM}
\newline {\small Structural Similarity Index}
&
$\mathrm{SSIM}(I,\hat{I}) = \dfrac{(2\mu_{I}\mu_{\hat{I}} + c_1)(2\sigma_{I\hat{I}} + c_2)}{(\mu_{I}^{2} + \mu_{\hat{I}}^{2} + c_1)(\sigma_{I}^{2} + \sigma_{\hat{I}}^{2} + c_2)}$
\newline\smallskip
{\small where $c_1$ and $c_2$ are stability constants. Captures luminance, contrast, and structural similarity jointly.}
&
\textbf{Closer to 1 is better.} \newline Range $[-1, 1]$; a score of $1$ indicates perfect structural agreement with the reference image. \\

\midrule
\textbf{DSC}
\newline {\small Dice Similarity Coefficient}
&
$\mathrm{DSC} =  \dfrac{2 |A \cap B|}{|A| + |B|} $
\newline\smallskip
{\small where $A$ and $B$ are the number of elements in each set.}
&
\textbf{Closer to 1 is better.} \newline A score of $1$ indicates a perfect overlap between a predicted segmentation mask and the ground truth. \\

\bottomrule
\end{tabular}
\end{table}

Based on a detailed analysis of the reviewed literature, four primary application categories were identified: (i) artifact removal, (ii) data augmentation, (iii) pseudo-healthy tissue inpainting, and (iv) other applications. In the following subsections, the works reviewed from each of these categories are discussed in more detail.

\subsection{Artifact Removal}

Artifact removal is a critical task in medical imaging, as artifacts can significantly degrade image quality and compromise diagnostic reliability. These artifacts may arise from several sources, including interference during image acquisition, patient motion, hardware limitations, metallic implants, incomplete projection data, and other acquisition-related inconsistencies. In this section, 18 studies employing diffusion-based inpainting approaches for artifact removal in medical imaging were identified and are summarized in Table \ref{tab:my-table2}.

{\footnotesize
\setlength\LTleft{0pt}
\setlength\LTright{0pt}

\begin{longtable}{p{0.18\textwidth}
                  p{0.15\textwidth}
                  p{0.05\textwidth}
                  p{0.08\textwidth}
                  p{0.25\textwidth}
                  p{0.15\textwidth}}

\caption{Summary of Reviewed Works on Artifact Removal Inpainting.}
\label{tab:my-table2}\\

\toprule
Artifact Type &
Article &
Year &
Category &
Image Distribution &
Results \\
\midrule
\endfirsthead

\toprule
Artifact Type &
Article &
Year &
Category &
Image Distribution &
Results \\
\midrule
\endhead

\midrule
\multicolumn{6}{r}{Continued on next page} \\
\endfoot

\bottomrule
\endlastfoot

Metal Artifact &
Tong et al.~\cite{Tong2023} &
2023 &
DDPM &
1,000 CT Sinograms from the DeepLesion Dataset &
\begin{tabular}[t]{@{}l@{}}
SSIM = 0.961\\
PSNR = 33.440
\end{tabular} \\
\\

&
Mei et al.~\cite{Mei2023} &
2023 &
SGM &
50 CBCT Volumes from the SICAS Medical Image Repository &
\begin{tabular}[t]{@{}l@{}}
MAE = 0.069\\
PSNR = 43.070
\end{tabular} \\
\\

&
Karageorgos et al.~\cite{Karageorgos2024} &
2024 &
DDPM &
11,560 CT Sinograms from the DeepLesion and UCLH Stroke EIT Datasets &
\begin{tabular}[t]{@{}l@{}}
SSIM = 0.964\\
PSNR = 46.160
\end{tabular} \\
\\

&
Wu et al.~\cite{Wu2025} &
2025 &
DDPM &
4,000 CT Images from the DeepLesion Dataset &
\begin{tabular}[t]{@{}l@{}}
SSIM = 0.966\\
PSNR = 33.440
\end{tabular} \\
\\

&
Wu et al.~\cite{WuZhong2025} &
2025 &
DDPM &
10 CBCT Volumes from the 2016 NIH-AAPM LDCT Grand Challenge Dataset &
\begin{tabular}[t]{@{}l@{}}
SSIM = 0.914\\
PSNR = 35.010
\end{tabular} \\
\\

&
Schaub et al.~\cite{Schaub2026} &
2025 &
SGM/NCSN &
10 CBCT Volumes from a Private Dataset &
\begin{tabular}[t]{@{}l@{}}
SSIM = 0.985\\
PSNR = 50.388
\end{tabular} \\
\midrule
Missing Angular Data &
Koo~\cite{Koo2023} &
2023 &
SDE &
3 Phantom CT Sinograms &
\begin{tabular}[t]{@{}l@{}}
PSNR = 42.215\\
SSIM = 0.930
\end{tabular} \\
\\

&
Guo et al.~\cite{Guo2025} &
2025 &
SDE &
\begin{tabular}[t]{@{}l@{}}
21,100 CT Sinograms from\\
the C4KC-KiTS Dataset
\end{tabular} &
\begin{tabular}[t]{@{}l@{}}
PSNR = 37.510\\
SSIM = 0.968\\
LPIPS = 0.012
\end{tabular} \\
\midrule

  Jewelry Obstruction &
Ji and He~\cite{Ji2023} & 2023&
  DDPM & 
  3,247 Bone Scans Images from the BS-80K Dataset & \begin{tabular}[t]{@{}l@{}} PSNR = 30.365 \\ SSIM = 0.882 \end{tabular} \\
   \midrule
Black Shadows & 
Dong et al.~\cite{Dong2024} & 2024& 
  DDPM & 
  9,416 Ophthalmic Images from  the Biobank, DRIVE, Cataract, MMAC and Messidor Datasets & \begin{tabular}[t]{@{}l@{}}PSNR = 37.470 \\ SSIM = 0.950\end{tabular} \\ \midrule
Tissue Folds and Bubbles &
Wang et al. \cite{Wang2024_1} &
  2024 &
  DDPM &
100,718 Histopathologic Images from the NCT-CRC-HE-100K and the CRC-VAL-HE-7K Datasets &
  ACC = 0.938 \\ \midrule

\begin{tabular}[t]{@{}l@{}}10 Types of Artifacts  \\ such as Dark Spots,\\ Blood Cells, Overlaps\\  and Folding\end{tabular} &
Fuchs et al.~\cite{Fuchs2024} & 2024 & DDPM &   13,106 Histopathologic Images from the Breast Cancer Semantic Segmentation Dataset &\begin{tabular}[t]{@{}l@{}} FID, PSNR: \\ Dark Spots: 70.3, 18.9 \\ Blood Cells: 33.8, 23.7  \\ Overlaps: 26.7, 23.5 \\ Folding: 35.4, 20.4 \end{tabular}
   \\ \midrule
Wide Saturation &
Ji et al.~\cite{Ji2024} & 2024
   &
  DDPM & 
  84,484 OCT Images from the Retinal OCT Dataset & 
  \begin{tabular}[t]{@{}l@{}}  SSIM = 0.744  \\ PSNR = 25.596 \\ ACC = 0.926\end{tabular}
   \\ \midrule
\begin{tabular}[t]{@{}l@{}} Missing Sections, \\ Tissue Tears and \\ Staining Errors\end{tabular} &
Kropp et al.~\cite{Kropp2024} & 2024 &
  DDPM & 
  30,000 Cellular Brain Images from the BigBrain Model &
   \begin{tabular}[t]{@{}l@{}}  FCD = 0.004 \end{tabular}
   \\   
Specular Reflection &
Tamang et al. \cite{Tamang2025} & 2025 & LDM &
  913 Colonoscopy Images from International Agency for Research on Cancer Dataset &
  SSIM = 0.850 \\
  \midrule
\begin{tabular}[t]{@{}l@{}}Motion and Obscured \\ Microvascular Artifacts\end{tabular} &
Xu et al. \cite{Xu2026} &
  2025 &
  DDPM &
  3,200 B-scans from a Private OCT Angiography Dataset &
  \begin{tabular}[t]{@{}l@{}}PSNR = 52.590\\ SSIM = 0.989\end{tabular} \\
  \midrule
Endoscopic Artifacts &
Yu et al. \cite{yu2025endoscopic} & 2025 & LDM
   & 
612, 1,000, 380 and 196 Colonoscopy Images from the CVC-ClinicDB, Kvasir,  CVC-ColonDB and ETIS Datasets and 195 Dental Endoscopy Images from a Private Dataset &
   \begin{tabular}[t]{@{}l@{}} DSC: \\
CVC-ClinicDB = 0.950\\
Kvasir-SEG = 0.933 \\
CVC-ColonDB = 0.815 \\
ETIS = 0.810 \\ 
Dental Dataset = 0.961
\end{tabular}
   \\
  \midrule 
  \begin{tabular}[t]{@{}l@{}}Pressure Line Artifacts\end{tabular} &
  Qiao and Hou \cite{Qiao2025} & 2025  & DDPM &
  1,681 Corneal Confocal Microscopy Images from 3 Private Datasets &
  \begin{tabular}[t]{@{}l@{}}SSIM = 0.984 \\ PSNR = 17.680 \\ MAE = 14.80\end{tabular} \\

\end{longtable}
}

The type of artifact addressed is closely related to both the imaging modality and the underlying acquisition problem. Across the reviewed studies, 12 distinct categories of artifacts were identified. Among these, metal artifacts in CT imaging were the most extensively investigated, appearing in six studies, while missing angular data in CT reconstruction was addressed in two works. Additional types included motion artifacts, missing image sections, jewelry-induced obstructions, dark spot artifacts, and modality-specific degradations associated with MRI, CT, and cone-beam CT (CBCT) imaging.

DDPMs were the most commonly adopted architectural family, reflecting their flexibility and strong reconstruction performance across diverse imaging tasks. Their widespread use also highlights the relative simplicity of adapting DDPM-based frameworks to different artifact removal scenarios. SGMs were employed in two studies focused on CBCT metal artifact reduction; however, these works were limited by relatively small datasets, which may restrict the generalizability of the reported results \cite{WuZhong2025,Schaub2026}. Similarly, SDE-based approaches were primarily applied to sinogram-domain CT reconstruction tasks involving missing angular data. A sinogram is a representation of the raw projection data used in tomographic imaging, in which each image-space point corresponds to a sinusoidal trajectory in the projection domain. Although these methods demonstrated promising performance, some studies still relied heavily on phantom-based datasets, limiting clinical validation \cite{Koo2023}.

Two studies specifically explored SDE-based frameworks for recovering missing angular projection data, using substantially different datasets and experimental configurations \cite{Koo2023,Guo2025}. Notably, the more recent work demonstrated significant improvements in both dataset scale and reconstruction performance, suggesting a rapid maturation of diffusion-based sinogram completion techniques.

Due to the broad diversity of artifact types, imaging modalities, and evaluation protocols, it is difficult to identify a single approach that consistently outperforms all others across every scenario. Nevertheless, several studies stand out because of their robust experimental design, large-scale datasets, and strong quantitative performance.
Among these, the work of Karageorgos et al.~\cite{Karageorgos2024} proposed a DDPM-based framework for inpainting missing sinogram data to improve metal artifact reduction (MAR). A notable characteristic of this approach is that the diffusion model was trained unconditionally, without requiring explicit information regarding the location or type of metal implants. This design choice potentially enhances the generalization capability of the method across different implant configurations in comparison with conditionally trained approaches. The method achieved strong quantitative performance, particularly considering the large-scale dataset used for evaluation.

Another notable contribution was presented by Guo et al.~\cite{Guo2025}, who introduced a diffusion framework based on mean-reverting SDEs to reconstruct missing angular data directly at the projection level. By combining diffusion distillation techniques with constraints derived from the pseudo-inverse of the inpainting matrix, the authors significantly accelerated the reconstruction process, reducing sampling to a near single-step procedure. This substantially improved computational efficiency while maintaining accurate and high-quality sinogram completion results.

\subsection{Inpainting for Data Augmentation to Improve Performance of Downstream Tasks}

Image inpainting has also been applied to another important challenge in machine learning: class imbalance in medical imaging datasets. In many clinical applications, certain pathological conditions or rare disease patterns are significantly underrepresented, which can negatively affect the training and generalization capability of deep learning models \cite{ghosh2024class}. In this context, image inpainting techniques can be employed as a data augmentation strategy to synthetically generate anatomically plausible variations of underrepresented classes, thereby improving dataset balance and diversity. Although this task does not fall in the category of what is typically called inpainting, we decided to include it in the review as it involves generating new data, rather than simply processing (e.g., restoring, segmenting, denoising) existing data \cite{morao2025data}. This augmentation process can enhance the performance of downstream tasks such as lesion detection, disease classification, and medical image segmentation. 

In total, 16 studies applying diffusion-based inpainting techniques for data augmentation purposes were identified in this review, and a detailed summary of these works is presented in Table~\ref{tab:my-table3}.

{\footnotesize
\setlength\LTleft{0pt}
\setlength\LTright{0pt}

\begin{longtable}{p{0.15\textwidth}
                  p{0.05\textwidth}
                  p{0.05\textwidth}
                  p{0.20\textwidth}
                  p{0.25\textwidth}
                  p{0.15\textwidth}}

\caption{Summary of Reviewed Works on Inpainting for Data Augmentation.}
\label{tab:my-table3} \\

\toprule
\multicolumn{1}{c}{Article} &
Year &
\multicolumn{1}{c}{Category} &
\multicolumn{1}{c}{Objective} &
\multicolumn{1}{c}{Dataset} &
Results \\
\midrule
\endfirsthead

\toprule
\multicolumn{1}{c}{Article} &
Year &
\multicolumn{1}{c}{Category} &
\multicolumn{1}{c}{Objective} &
\multicolumn{1}{c}{Dataset} &
Results \\
\midrule
\endhead

\midrule
\multicolumn{6}{r}{Continued on next page} \\
\endfoot

\bottomrule
\endlastfoot

Kadi et al.~\cite{Kadi2023} & 2023 &
LDM &
\begin{tabular}[t]{@{}l@{}}Oral Rare Disease\\ Detection\end{tabular} &
156 Oral X-Rays from a Private Dataset &
\begin{tabular}[t]{@{}l@{}}AP = 0.76\end{tabular} \\ 
\\
Ma et al.~\cite{Ma2024} & 2024 &
LDM &
Polyp Segmentation &
51,241 Colonoscopy Images from the Kvasir-SEG, ETIS-LaribPolypDB,SUN-SEG and CVC-EndoSceneStill Datasets &
\begin{tabular}[t]{@{}l@{}}DSC = 0.846\\ IoU = 0.778\end{tabular} \\
\\
Montoya et al.~\cite{Montoya2024} & 2024 &
LDM &
Mass Lesion Synthesis &
55,302 Mammography Images from the OMI-H and the VinDr-Mammo Datasets &
MS-SSIM = 0.37 \\
\\
Montoya et al.~\cite{Montoya2025} & 2024 &
LDM &
Lesion Detection &
10,036 Mammography Images from CDD-CESM and Private Datasets &
\begin{tabular}[t]{@{}l@{}}Sensitivity = 0.76\\ AUROC = 0.73\end{tabular} \\
\\
Wang et al.~\cite{Wang2024} & 2024 &
DDPM &
Dental Segmentation &
29,112 Panoramic Radiographs Images from a Private Dataset &
\begin{tabular}[t]{@{}l@{}}DSC:\\ Instance = 0.936\\ Enamel = 0.791\\ Dentin = 0.856\\ Pulp = 0.703\end{tabular} \\
\\
Graf et al.~\cite{Graf24a} & 2024 &
DDPM &
Spine Segmentation &
416 Spine MRI Images from a Private Dataset &
DSC = 0.718 \\
\\
Yang et al.~\cite{Yang2025} & 2025 &
LDM &
Lesion Classification &
\begin{tabular}[t]{@{}l@{}}10,000 Dermatological Images\\ from HAM10000 Dataset\end{tabular} &
ACC = 0.780 \\
\\
Zhang et al.~\cite{Zhang2024_1} & 2025 &
DDPM &
Lesion Segmentation &
26,800 T1w and 26,000 FLAIR MRI Slices from a Private Dataset and 20 T1w and FLAIR Volumes from the 2015 ISBI Challenge Dataset &
\begin{tabular}[t]{@{}l@{}}DSC:\\ Brainstem = 0.971\\ Gray Matter = 0.949\\ Ventricles = 0.932\\ White Matter = 0.915\\ Thalamus = 0.936\end{tabular} \\
\\
Zhang et al.~\cite{Zhang2025} & 2025 &
DDPM &
\begin{tabular}[t]{@{}l@{}}Lung Node and Cardiac\\ Lesion Segmentation\end{tabular} &
\begin{tabular}[t]{@{}l@{}}1,077 Histopathologic Images from\\ the LIDC and Emidec Datasets\end{tabular} &
\begin{tabular}[t]{@{}l@{}}LIDC, Emidec:\\ DSC = 0.834, 0.713\end{tabular} \\
\\
Wang et al.~\cite{Wang2025} & 2025 &
DDPM &
\begin{tabular}[t]{@{}l@{}}Rare-type Nuclei\\ Segmentation and \\Classification\end{tabular} &
50k Histopathological Images from TCGA and 309 Images from CoNSeP, GLySAC, MoNuSAC Datasets &
\begin{tabular}[t]{@{}l@{}}DSC, ACC:\\ CoNSeP = 0.832, 0.744\\ GLySAC = 0.822, 0.842\\ MoNuSAC = 0.782, 0.844\end{tabular} \\
\\
He et al.~\cite{He2026} & 2025 &
LDM &
\begin{tabular}[t]{@{}l@{}}Vertebral Disease\\ Classification\end{tabular} &
3,207 Vertebra MRI Images from VerTumor1200 and RSNA 2024 Lumbar Spine Degenerative Classification Dataset &
\begin{tabular}[t]{@{}l@{}}FID = 2.875\\ PSNR = 10.845\\ SSIM = 0.669\end{tabular} \\
\\
Nazir et al.~\cite{nazir2025diffusion} & 2025 &
LDM &
Polyp Segmentation &
1,612 Colonoscopy Images from CVC-ClinicDB and Kvasir-SEG and 1,200 Fundus Images from REFUGE2 Dataset &
\begin{tabular}[t]{@{}l@{}}DSC:\\ CVC-ClinicDB = 0.964\\ Kvasir-SEG = 0.956\\ REFUGE2 = 0.902\end{tabular} \\
\\
Lei et al.~\cite{Lei2025} & 2025 &
LDM &
\begin{tabular}[t]{@{}l@{}}Multi-Organ Lesion\\ Segmentation\end{tabular} &
\begin{tabular}[t]{@{}l@{}}12,372 CT volumes from KiTS23, MSD,\\ INSTANCE22 and Private Datasets\end{tabular} &
DSC = 0.701 \\
\\
Prochazka et al.~\cite{Prochazka2026} & 2025 &
LDM &
Nodule Segmentation &
4,032 Ultrasound Images from TN3K, TDID and TUCC Datasets &
\begin{tabular}[t]{@{}l@{}}DSC, IoU, Accuracy:\\ TN3K: 0.814, 0.726, 0.970\\ TDID: 0.592, 0.758, 0.942\\ TUCC: 0.579, 0.474, 0.948\end{tabular} \\
\\
Liu et al.~\cite{Liu2026} & 2025 &
LDM &
Tumor Segmentation &
632 Histopathologic Images from Ductal Carcinoma in Situ, Pan-tumor Canine Cutaneous Cancer Histology and CAMELYON16 Datasets  &
IoU = 0.777 \\
\\
Kim and Park~\cite{Kim2026} & 2026 &
LDM &
\begin{tabular}[t]{@{}l@{}}Tumour and Normal\\ Brain Tissue Synthesis\end{tabular} &
1,219 Brain MRI Images from BraTS Dataset &
\begin{tabular}[t]{@{}l@{}}Tumor, Normal:\\ FID = 13.27, 13.35\\ LPIPS = 0.088, 0.081\end{tabular} \\

\end{longtable}
}

Ten of reviewed methodologies focused primarily on segmentation tasks, four addressed image classification, two targeted detection performance, and two investigated synthesis applications. This distribution highlights the strong relevance of diffusion-based augmentation techniques for improving segmentation tasks, where class imbalance and limited annotated data remain an area that requires further research. In particular, segmentation models typically require large quantities of pixel-level annotations, which are expensive, time-consuming, and highly dependent on expert clinical knowledge.

Regarding architectural choices, eleven out of the sixteen reviewed studies employed LDM-based approaches, while the remaining approaches were based on DDPMs. The predominance of LDM-based methods suggests a growing preference for latent-space diffusion techniques in data augmentation applications, primarily due to their improved computational efficiency and their ability to generate high-quality and semantically consistent synthetic samples. These characteristics are particularly important in scenarios requiring the realistic transfer or synthesis of pathological patterns to augment underrepresented disease classes.

The reviewed approaches were applied across multiple imaging modalities, including histopathological imaging, MRI, and mammography. Among these, histopathological images represented the most frequently explored modality, likely due to the scarcity of annotated samples for rare pathological conditions in publicly available datasets. In mammography, diffusion-based inpainting methods were primarily investigated for lesion synthesis and lesion detection enhancement, demonstrating promising potential for improving computer-aided diagnosis systems.

Most studies focused on a single organ, lesion type, or pathology at a time, reflecting the highly specialized nature of current medical image augmentation pipelines. Only one study explored a multi-organ augmentation framework, highlighting a significant research gap in the development of generalized diffusion-based inpainting approaches capable of handling multiple anatomical structures and disease categories simultaneously.

\subsection{Inpainting Pseudo-Healthy Tissue}

Pseudo-healthy tissue inpainting is a preprocessing technique aimed at removing lesions, tumors, or other pathological anomalies from medical images, particularly MRI scans, and replacing them with synthetically reconstructed healthy tissue \cite{Durrer2026}. This approach is especially valuable because it preserves patient-specific anatomical structures while mitigating the influence of pathological regions, which is crucial for data augmentation, clinical disease diagnosis, and understanding pathology-induced changes \cite{liu2024lesion}. Furthermore, pseudo-healthy reconstruction can improve the robustness of downstream tasks, including segmentation, registration, and treatment planning, by reducing the bias introduced by abnormal tissue regions.

A variety of methods have been proposed for this problem, and this review identified twelve studies employing diffusion-based approaches to address it. A detailed summary of these works is provided in Table~\ref{tab:my-table1}.

{\footnotesize
\setlength\LTleft{0pt}
\setlength\LTright{0pt}
\begin{longtable}{p{0.15\textwidth}
                  p{0.05\textwidth}
                  p{0.08\textwidth}
                  p{0.18\textwidth}
                  p{0.25\textwidth}
                  p{0.15\textwidth}}

\caption{Summary of Reviewed Works on Pseudo-Healthy Tissue Inpainting.}
\label{tab:my-table1}\\

\toprule
Article &
Year &
Category &
Disease Type &
Image Distribution &
Results \\
\midrule
\endfirsthead

\toprule
Article &
Year &
Category &
Disease Type &
Image Distribution &
Results \\
\midrule
\endhead

\midrule
\multicolumn{6}{r}{Continued on next page} \\
\endfoot

\bottomrule
\endlastfoot

Durrer et al.~\cite{Durrer2023} & 2023&
   DDPM & 
   Brain Tumor
   & 
   1470 T1 MRI Images from the BraTS Dataset & \begin{tabular}[t]{@{}l@{}} SSIM = 0.827 \\ MSE = 0.012 \\ PSNR = 20.495 \end{tabular}
   \\
   \\
Durrer et al.~\cite{Durrer2025} & 2025
   & 
  DDPM & 
  Brain Tumor
   & 
   1,819 MRI images from the BraTS Dataset & \begin{tabular}[t]{@{}l@{}} SSIM = 0.853 \\ MSE = 0.103 \\ PSNR=20.926 \end{tabular}
   \\
   \\
Jiang et al.~\cite{Jiang2025} & 2025   &
  LDM & 
  \begin{tabular}[t]{@{}l@{}}Pancreas, Lung, \\ Liver, Colon, \\ Hepatic Vessel \\ and Kidney Tumor\end{tabular}
   &  3,933 MRI and CT images from KiTS23, MSD, BraTS and Private Datasets 
   & 
  \begin{tabular}[t]{@{}l@{}} DSC: \\ Pancreas=0.436 \\ Lung = 0.429 \\ Liver = 0.632 \\ Colon = 0.387 \\ Hepatic = 0.628 \\ Kidney = 0.598 \end{tabular}
  \\
  \\
Pollak et al. \cite{Pollak2025} & 2025 & LDM &
  Brain Tumors, Cavities and Abnormalities &
  2,727 MRI volumes from HCP, RS, ABIDE, ADNI, IXI, LA5C, MBB, MIRIAD, OASIS, UPENN-GBM, BTC, UKB and Private Dataset &
  \begin{tabular}[t]{@{}l@{}}PSNR = 29.640\\ SSIM = 0.720\\ DSC = 0.949\end{tabular} \\ 
  \\
  Wehrli et al. \cite{Wehrli2025} &
  2025 &
  DDPM &
  \begin{tabular}[t]{@{}c@{}}Trochlear \\ Dysplasia \end{tabular} &
  1,560 MRI Images from the fastMRI and Private datasets  &
  \begin{tabular}[t]{@{}l@{}}MSE = 0.025\\ PSNR = 16.345\\ SSIM = 0.065\end{tabular} \\
  \\
  Tao et al. \cite{Tao2025} &
  2025 &
  DDPM &
  Brain Tumor &
  1,257 MRI volumes  from the BraTS Dataset &
  \begin{tabular}[t]{@{}l@{}}PSNR = 20.059\\ SSIM = 0.804\\ MSE = 0.012\end{tabular} \\
  \\
  \\
  Ferreira et al.~\cite{ferreira2025achieving} & 2025 &
   DDPM & 
   Brain Tumor & 
    1,450 MRI images from the BraTS 2023 Glioma dataset & 
     \begin{tabular}[t]{@{}l@{}}MSE = 0.005 \\ PSNR = 24.529 \\SSIM = 0.867\end{tabular}
   \\
   \\
Zhu et al. \cite{zhu2025patchstructdiffusion} & 2025 &
   DDPM  & 
   Brain Tumor &
   7,667 T1W MRI images from ADNI Dataset &
   \begin{tabular}[t]{@{}l@{}} SSIM = 0.9670 \\ PSNR = 29.6436 \\ MAE = 0.0106 \end{tabular}
   \\ 
   \\
Dai et al.~\cite{dai2025context} & 2025 &
   DDPM & 
   Brain Tumor & 
   1,471 MRI images from the BraTS 2025 Local Inpainting Challenge Dataset  & 
     \begin{tabular}[t]{@{}l@{}}SSIM = 0.721 \\ PSNR = 17.700 \\MSE = 0.021\end{tabular}
   \\
   \\
Koch et al. \cite{koch2025local} & 2025 &
   Transformer DDPM & 
   Brain Tumor & 
   1,501 MRI images from the BraTS 2025 Local Inpainting Challenge Dataset & 
     \begin{tabular}[t]{@{}l@{}}PSNR = 21.469 \\SSIM = 0.837\end{tabular}
   \\
   \\
  Kwark et al.~\cite{Kwark2026} & 2026 &
  DDPM 
   & Brain Tumor
   &
  2,312 MRI volumes from the BraTS Dataset and  the HCP Dataset & 
  \begin{tabular}[t]{@{}l@{}} BraTS: \\ SSIM = 0.805 \\ MSE = 0.008 \\ PSNR = 22.039 \\ HCP: \\ SSIM = 0.598 \\ MSE = 0.017 \\ PSNR = 17.845 \end{tabular} \\
   \\
     Durrer et al.~\cite{Durrer2026} & 2026
   &
  DDPM & 
  Brain Tumor
   & 1,820 MRI volumes from BraTS Dataset and Private Datasets
  & \begin{tabular}[t]{@{}l@{}} SSIM = 0.857 \\ MSE = 0.008 \\ PSNR = 22.260 \end{tabular}
  \\ 
\end{longtable}

}

Among the reviewed methods, DDPMs constitute the dominant architectural choice, reflecting a clear evolution from early 2D approaches toward fully 3D diffusion-based frameworks. This transition is particularly important in medical imaging, as 3D models are better suited to capture volumetric anatomical consistency and inter-slice spatial dependencies. In contrast, only two studies explored LDMs, which reflects the relatively recent adoption of latent-space diffusion techniques in medical imaging applications. Interestingly, within this small subset of LDM-based approaches, one study employed the traditional convolutional U-Net as its backbone, while the other distinguished itself by introducing a DiT~\cite{koch2025local}.

Most of the identified studies focus on brain MRI data. This predominance can be attributed both to the clinical relevance of pseudo-healthy reconstruction in neuroimaging applications, to the widespread availability of publicly accessible brain imaging datasets, and the existence of public challenges that encourage the development and benchmarking of these approaches. In comparison, only a limited number of studies investigated applications involving multi-organ imaging or specific conditions such as trochlear dysplasia.

Regarding datasets, most studies relied primarily on publicly available benchmarks, with the BraTS dataset emerging as the most frequently used resource. Several works additionally combined public and private datasets to construct training, validation, and testing cohorts ranging from approximately 1,000 to 4,000 images. The use of heterogeneous datasets generally contributes to improved model robustness, enhanced generalization capability, and more reliable evaluation of reconstruction performance across diverse clinical scenarios.

Among the reviewed methods, the work of Durrer et al.~\cite{Durrer2026} appears to represent one of the most effective approaches for pseudo-healthy tissue inpainting. The authors proposed \textit{fastWDM3D}, a wavelet-based diffusion framework operating in the discrete wavelet domain. By reducing the number of diffusion sampling steps to as few as two while maintaining reconstruction quality, the method substantially improves sampling efficiency compared with conventional diffusion approaches. 



\subsection{Other Applications}
Additionally, several other applications of diffusion-based image inpainting were identified in the reviewed literature, including data imputation, anomaly detection, 3D image generation from 2D inputs, and cranial implant design. A summary of these nine studies is presented in Table~\ref{tab:new-my-table}.

These applications are primarily focused on head CT and MRI imaging, reflecting the strong clinical relevance of neuroimaging tasks in diffusion-based reconstruction research. In contrast to the more standardized applications discussed previously, these methods often involve highly specialized implementation strategies tailored to specific clinical or anatomical requirements.  Overall, these emerging applications highlight the versatility of diffusion-based inpainting frameworks and demonstrate their potential beyond traditional image restoration tasks.

{\footnotesize
\setlength\LTleft{0pt}
\setlength\LTright{0pt}
\begin{longtable}{p{0.15\textwidth}
                  p{0.15\textwidth}
                  p{0.05\textwidth}
                  p{0.08\textwidth}
                  p{0.22\textwidth}
                  p{0.2\textwidth}}
                  
\caption{Summary of Reviewed Works on Inpainting for Other Applications.}
\label{tab:new-my-table}\\

\toprule
Objective &
Article &
Year &
Category &
Image Distribution &
Results \\
\midrule
\endfirsthead

\toprule
Objective &
Article &
Year &
Category &
Image Distribution &
Results \\
\midrule
\endhead

\midrule
\multicolumn{6}{r}{Continued on next page} \\
\endfoot

\bottomrule
\endlastfoot

\begin{tabular}[t]{@{}l@{}}Inpainting Temporal or \\Longitudinal Gaps\end{tabular} &
Bae et al.~\cite{Bae2024} & 2024 &
  DDPM &
  551 Cerebral CT Perfuration images from ISLES2018 Challenge, UniToBrain and Private Datasets &
  \begin{tabular}[t]{@{}l@{}}SSIM = 0.915\\ PSNR = 32.002\\  FID = 0.001\end{tabular} \\
  \\
  &
Guo et al.~\cite{GuoTao2024} & 2024 &
  DDPM &
  655 T1w Brain MRI from the Baby Connectome Project & \begin{tabular}[t]{@{}l@{}} PSNR = 24.150 \\ SSIM = 0.810 \end{tabular} \\
   \\
  &
Zhu et al. \cite{Zhu2024} &
  2024 &
  DDPM &
  655 T1w Brain MRI from the Baby Connectome Project &
  \begin{tabular}[t]{@{}l@{}}PSNR = 25.520\\ SSIM = 0.845\\ DSC = 0.650\end{tabular} \\ 

&
Tao et al. \cite{tao2026trustworthy} & 2026 &
   DDPM & 
   2,535 T1w  Brain MRIs from the OASIS-3 dataset
  &
  \begin{tabular}[t]{@{}l@{}}PSNR = 25.52\\ SSIM = 0.84 \end{tabular}
   \\
   \\
&
You et al. \cite{you2025fb} & 2025 & LDM
   & 275 4D MRI images from the ACDC Cardiac and 4D-Lung Datasets
   & \begin{tabular}[t]{@{}l@{}} FID = 32.7 \\ PSNR = 26.18 \\LPIPS = 2.287  \end{tabular}
   \\
  
  \midrule
  
\begin{tabular}[t]{@{}l@{}}Inpainting-Based \\ Anomaly Detection \end{tabular} &
Olsen et al. \cite{Olsen2025} & 2024 & DDPM &
 14,068 Ultrasound Images from the Danish National Fetal Ultrasound Screening Dataset &
  \begin{tabular}[t]{@{}l@{}}FID = 48.390\\ SSIM = 0.989\\ AUROC = 0.620\end{tabular} \\
  \\
  &
Shi et al. \cite{Shi2025} & 2025 & DDPM &
  11,864 liver CT slices from LiTS and IRCAD Datasets &
  \begin{tabular}[t]{@{}l@{}}LiTS, IRCAD\\ AUC = 0.881, 0.764\\ AP = 62.24, 57.29\\ DSC = 0.621, 0.585\end{tabular} \\ \midrule
\begin{tabular}[t]{@{}l@{}}Generating 3D \\ from 2D images\end{tabular} &
Jeong et al.~\cite{Jeong2023} & 2023
   &
  SDE & 
  1,000 Brain CT Slices from Private Dataset & \begin{tabular}[t]{@{}l@{}} FID-Ax = 14.993 \\ FID-Cor = 19.188 \\ FID-Sag = 19.698 \end{tabular} \\
   \\
   &
Zuo et al. \cite{Zuo2025} &
  2025 &
  LDM &
  30 Abdominal CT images from Private Dataset &
  \begin{tabular}[t]{@{}l@{}}Error Rate (viscera fat) = 26.3\\ Error Rate (muscle) = 15.2\end{tabular} \\ \midrule
Design 3D Cranial Implants &
Liu et al. \cite{Liu2024} & 2024 &
  LDM &
  364 Head CT volumes from SkullFix, SkullBreak and Private Database &
  DSC = 0.924 \\ \midrule
Stain Imputation &
Li et al.~\cite{Li2025} & 2025 &
  DDPM & 
  83 Multiplex Immunofluorescence Images from a Private Dataset & \begin{tabular}[t]{@{}l@{}} MAE = 9.78 \\ SSIM = 0.579 \\ PSNR = 22.79  \end{tabular}
  \\ \midrule

Synthesis of Progression Sequences for Diabetic Foot Ulcers & 
Hahne et al.~\cite{hahne2026ai} & 2026 & 
   DDPM & 

   Private Dataset
   &
   None
   \\
\end{longtable}%
}

\section{Discussion}
\label{sec:discussion}

In this section, the main findings derived from the reviewed studies are discussed, including datasets, methodological trends, and diffusion model categories.

\subsection{Image Datasets}

The analysis of the reviewed studies reveals several important trends regarding dataset usage in diffusion-based medical image inpainting research. One of the most notable observations is the limited reuse of standardized public benchmarks across the literature. Among all reviewed datasets, only the BraTS and DeepLesion datasets were employed in more than one study. The limited overlap in dataset usage suggests a lack of standardized evaluation protocols within the field, complicating comparisons between methods and limiting the reproducibility and generalizability of the results.

In addition to the limited use of common public datasets, approximately 16 studies relied partially or entirely on private clinical datasets. Although private datasets may correspond to highly specialized clinical scenarios and institution-specific imaging protocols, their restricted accessibility poses important challenges for reproducibility, benchmarking, and independent validation. This reliance on proprietary data also limits the ability of the research community to perform fair cross-method comparisons and may hinder the development of standardized evaluation frameworks for medical image inpainting.

Another important observation concerns dataset scale. Most reviewed studies used datasets containing approximately 1,000 to 3,000 images distributed across training, validation, and testing subsets. While such dataset sizes are relatively common in medical imaging research, they remain modest compared with the large-scale datasets typically required to fully exploit the generative capacity of modern diffusion models. This limitation is particularly relevant given the high complexity and parameter count of diffusion-based architectures, which generally benefit from extensive and diverse training data to achieve robust generalization performance. Consequently, many existing methods may be susceptible to overfitting, reduced robustness across institutions, or limited adaptability to unseen clinical scenarios.

The reviewed studies also revealed a strong predominance of neuroimaging and thoracic imaging applications, particularly involving brain MRI and CT imaging. Brain MRI emerged as the most extensively explored modality and anatomical region, largely due to the widespread availability of publicly accessible datasets such as BraTS and the high clinical relevance of tasks involving tumor pseudo-healthy tissue inpainting. Similarly, CT imaging was frequently investigated in artifact removal studies because of the clinical importance of correcting artifacts.

Other anatomical regions and imaging modalities were considerably less represented in the literature. Applications involving histopathological, mammography, and ophthalmological imaging appeared only in a limited number of studies. This imbalance likely reflects both the reduced availability of large annotated public datasets and the lower frequency of standardized benchmarks for these domains.

Overall, the current dataset landscape exhibit several important limitations, including limited dataset standardization, small sample sizes, and strong concentration on a few anatomical regions and imaging modalities.

\subsection{Methodology Categories}


The methodologies identified across the 60 reviewed studies reveal a strong application-specific trend in the development of diffusion-based medical image inpainting approaches. As discussed in the previous sections, most methods are designed for and evaluated on a single clinical objective, such as pseudo-healthy tissue reconstruction, artifact removal, or data augmentation, with limited investigation into their applicability across different medical imaging tasks. This specialization suggests that diffusion-based inpainting research remains highly task-oriented, with relatively little emphasis on cross-domain generalization or unified reconstruction frameworks.

This pattern is also reflected in the experimental settings adopted. Only the works of Jiang et al.~\cite{Jiang2025}, Fuchs et al.~\cite{Fuchs2024}, Kropp et al.~\cite{Kropp2024}, and Lei et al.~\cite{Lei2025} proposed frameworks capable of handling multiple pathological conditions, organs, or artifact categories. In contrast, the majority of approaches were validated under highly constrained conditions involving a single imaging modality, anatomical structure, or pathology type. Consequently, the current literature still lacks comprehensive studies evaluating the robustness and generalization capability of diffusion models across heterogeneous medical imaging tasks and clinical environments.

Among the identified application categories, artifact removal constitutes the largest group of reviewed studies. As discussed previously, this predominance is closely linked to the direct clinical importance of artifact correction in medical imaging workflows. In this context, diffusion-based inpainting methods have demonstrated strong potential for restoring anatomically consistent information while preserving clinically relevant structures across multiple imaging modalities.

Data augmentation emerged as the second most prominent application category. The growing interest in this area reflects the persistent challenges associated with limited dataset availability, class imbalance, and patient privacy constraints in healthcare imaging. As highlighted in the studies summarized in Table~\ref{tab:my-table3}, diffusion-based techniques can generate anatomically plausible synthetic samples that improve dataset diversity and enhance the performance of downstream tasks such as segmentation, classification, and detection. The predominance of LDM-based approaches in this category further suggests that latent-space diffusion provides practical advantages for scalable augmentation pipelines and computationally intensive learning tasks.

Pseudo-healthy tissue reconstruction represents another clinically relevant application domain, particularly in neuroimaging studies involving brain MRI. The reviewed works demonstrate that diffusion-based methods can effectively reconstruct healthy anatomical structures while preserving patient-specific information, enabling improved segmentation, registration, and pathology analysis. However, most studies remain restricted to brain imaging applications and relatively limited datasets, indicating that broader validation across additional organs and imaging modalities is still needed.

Overall, the reviewed literature demonstrates the versatility and strong reconstruction capability of diffusion-based inpainting methods across multiple healthcare applications. Nevertheless, the field remains fragmented, with most approaches optimized for highly specialized scenarios and limited evaluation of cross-task adaptability.

\subsection{Diffusion Model Categories}
Based on the proposed taxonomy in Section ~\ref{sec:background_knowledge} and from a theoretical perspective,  diffusion models can be broadly categorized into DDPMs, LDMs, SGMs, and SDE-based formulations. From a theoretical perspective, SDEs constitute the most general continuous-time framework for diffusion modeling, unifying both DDPMs and SGMs under a common probabilistic formulation.


Among the reviewed studies, DDPMs and LDMs clearly dominate the medical image inpainting literature. This dominance is likely associated with their implementation maturity, training stability, and strong reconstruction performance across diverse imaging tasks. In particular, the introduction of DDPMs by Ho et al.~\cite{Ho2020} significantly improved the practical applicability of diffusion models by reformulating diffusion processes into a discrete-time probabilistic framework that is easier to train and reproduce in real-world settings.

Despite their effectiveness, conventional DDPMs remain computationally demanding, as both training and inference typically require hundreds of iterative denoising steps. To mitigate this limitation, LDMs perform the diffusion process in compressed latent representations, substantially reducing computational and memory requirements while preserving reconstruction fidelity. This characteristic makes LDMs particularly attractive for large-scale applications and downstream tasks such as segmentation, classification, and data augmentation.

The reviewed literature suggests a clear preference for discrete and latent-space diffusion paradigms in medical image inpainting. DDPM-based methods were identified across all categories of the proposed taxonomy, whereas LDMs were more commonly associated with augmentation-oriented applications and computationally intensive scenarios. This trend likely reflects the practical advantages of latent-space representations when handling high-resolution and heterogeneous medical imaging datasets.

In contrast, SGM- and SDE-based approaches remain considerably underexplored in medical image inpainting. The reviewed studies employing SDE formulations mainly focused on specific reconstruction tasks, including 3D generation from 2D images~\cite{Jeong2023} and missing angular reconstruction~\cite{Guo2025,Koo2023}. Similarly, only a limited number of studies explicitly explored SGMs~\cite{Mei2023,Schaub2026}. This observation reveals several important research gaps in the current literature.

The limited adoption of SGM- and SDE-based methods reveals several important research opportunities. First, continuous-time diffusion formulations remain insufficiently explored for general medical image inpainting tasks, despite their potential for improved sampling flexibility and probabilistic modeling. Second, advanced score-based sampling strategies, such as predictor-corrector methods, annealed langevin dynamics, and adaptive SDE solvers, are rarely investigated in healthcare imaging pipelines. Finally, uncertainty-aware reconstruction remains largely unexplored, even though stochastic diffusion formulations naturally support the generation of multiple plausible reconstructions. This capability could be particularly valuable in clinical scenarios involving ambiguous anatomical structures or complex pathological regions, where uncertainty estimation may improve the interpretability and reliability of reconstructed medical images.

A promising future trend emerging in the application of LDM-based frameworks for medical image inpainting is the adoption of DiTs. While currently underrepresented in the literature, with only one reviewed study utilizing this approach, initial results are highly encouraging. For instance, when comparing the DiT-based method by 
Koch et al.\cite{koch2025local} and the standard U-Net-based diffusion approach by Dai et al.\cite{dai2025context} on the same BraTS2025 Challenge dataset, the DiT architecture demonstrates superior performance, outperforming the traditional model in both SSIM and PSNR metrics. This practical advantage in the medical domain aligns with the foundational findings of DiTs, which demonstrate that replacing the inductive bias of convolutional U-Nets with scalable Transformer backbones significantly improves the generation of high-fidelity representations~\cite{Peebles2023}.

\section{Case Study}
\label{sec:case_study}

To complement the systematic review with an empirical evaluation, we conducted a case study comparing diffusion-based image inpainting methods under a unified experimental protocol. The GitHub repositories associated with the selected studies were examined to identify publicly available implementations suitable for reproducible benchmarking. Based on this process, ten diffusion-based image inpainting models were selected.

The following sections describe the datasets, the experimental protocol adopted for generating missing regions (masks), the evaluation metrics, and the experimental results obtained in this case study.

\subsection{Overview}

We benchmark ten diffusion-based inpainting models under a controlled experiment that decouples the restoration model from the missing-data mechanism used to corrupt the input. These models include general-purpose 2D diffusion approaches (SDEdit, IR-SDE, DM$\_$Inpainting~\cite{Durrer2025}, ArtifactRestoration~\cite{Wang2024_1}, TPDM~\cite{Schaub2026}), 3D volumetric methods adapted to 2D inputs (fastWDM3D~\cite{Durrer2026}, NeuroLIT~\cite{Pollak2025}), and histopathology-specific models evaluated on both domains (HARP~\cite{Fuchs2024}, PathoPainter~\cite{Liu2026}, and NucleiMix~\cite{Wang2025}). The models were trained for $100$ epochs with learning rate $10^{-4}$.

For each image, the mask is generated from a deterministic seed, ensuring every model sees the same corruption for a given image. Moreover, scoring is region-restricted: before computing metrics, we copy the ground-truth pixels back into all observed locations ($\text{restored}[\lnot\text{mask}] = \text{original}[\lnot\text{mask}]$), so that the reported quality reflects only the inpainted region and is neither inflated nor penalized by model drift on known pixels. We report PSNR (dB) and SSIM (channel-wise, $7\times7$ window) between restoration and ground truth, aggregated as the mean and standard deviation per dataset, mask, and model.

Each experiment is a sweep over the Cartesian product of the datasets, mask strategies, and models: for every test image, we (i)~generate a deterministic binary mask, (ii)~remove the masked pixels to obtain the degraded observation, (iii)~run the model's inpainting routine, and (iv)~score the recovered region against ground truth. By holding the dataset and corruption process fixed while varying only the model, we attribute performance differences to either the restoration method or the nature of the missingness, rather than to confounded data pipelines.

\subsection{Data}
We evaluate on two complementary medical-imaging domains: TCGA histopathology patches (RGB, $224\times224$, stomach tissue, tumor/non-tumor) and BraTS brain-MRI axial mid-slices (RGB-encoded). The TCGA dataset has 1000 images, while BraTS has 251 images both in 2D format. Each dataset is split $80/20$ into train/test with a fixed seed; models are trained on the train split under the same mask strategy used at test time, and all metrics are reported on the held-out test split. All images are resized to $256\times256$ for model input. 

\subsection{Missing Data Generation Strategies}
\label{sec:masks}
The core experimental variable is the missingness is mechanism. Rather than a single ad-hoc corruption, we define three parametric mask families that span the classical taxonomy: Missing Completely at Random (MCAR), Missing at Random (MAR), and Missing Not at Random (MNAR), each motivated by a concrete imaging-artifact analog (see Table~\ref{tab:masks}). All strategies share the convention that $\text{mask}(i,j)=1$ marks a pixel as missing (to be inpainted) at location $(i,j)$. Moreover, let $r$ denote the target corruption rate (expected fraction of masked pixels).

\paragraph{1. Dead-Pixel Masking (MCAR)}
A fraction $r$ of pixel locations is drawn uniformly at random without replacement, independent of both spatial position and pixel value:
\begin{equation}
\Pr[\text{mask}(i,j)=1] = r, \qquad \forall\,(i,j).
\end{equation}
This reproduces sensor dead-pixel noise: isolated, spatially uncorrelated defects scattered across the field of view. It is the easiest regime for inpainting, as every missing pixel is surrounded by observed context, for moderate values of $r$.

\paragraph{2. Intensity-Extremes Masking (MNAR)}
The fraction $r$ of pixels with the most extreme intensity are masked, where intensity is the raw value for grayscale or the ITU-R BT.601 luminance for RGB ($0.299R + 0.587G + 0.114B$). Two modes are available: with \texttt{target="highest"}, the brightest pixels are removed (e.g., bright lesions in BraTS MRI); with \texttt{target="lowest"}, the darkest ones are deleted (e.g., deeply stained nuclei in TCGA). Here, missingness depends on the value of the missing pixel itself, which is the defining property of MNAR.

\paragraph{3. Sinusoidal-Stripe Masking (MAR)}
Missingness is governed by a spatially periodic probability field that depends only on the fully observed row coordinate:
\begin{equation}
p(i,j) = \operatorname{clip}\!\left(
  2r\,\sin^2\!\left(\frac{2\pi\,\mathrm{pos}}{\lambda} + \varphi\right),\,
  0,\, 1 \right),
\label{eq:sinstripe}
\end{equation}
with $\mathrm{pos}\in\{i,j\}$ selected by row, period $\lambda$ (pixels), and a per-image random phase $\varphi\sim\mathcal{U}[0,2\pi)$. The $\sin^2$ function has mean $\tfrac{1}{2}$ over a full period, so the factor of $2$ calibrates the expected masked fraction to $r$. Because missingness is a known function of position and is conditionally independent of pixel value, this is a clean MAR mechanism, and it directly emulates MRI k-space readout-line dropout, where whole bands of the image are attenuated by acquisition geometry rather than tissue content. The period controls band granularity (e.g.,\ $\lambda=32$\,px yields ${\sim}8$ bands across a $256$\,px image), interpolating between fine speckle and coarse structured occlusion.

\begin{table}[t]
\centering
\caption{The three missing-data generation strategies, spanning the classical missingness taxonomy. Corruption rate $r$ is held fixed across strategies so that difficulty reflects the mechanism of missingness.}
\label{tab:masks}
\begin{tabular}{@{}llccl@{}}
~\\\toprule
Strategy & Mechanism & Spatial Structure & Value-Dependent? & Artifact Analog \\
\midrule
Dead-pixel          & MCAR & None (random)   & No  & Sensor dead pixels \\
Intensity-extremes  & MNAR & Value-clustered & Yes & Saturated background / lesions \\
Sinusoidal-stripe   & MAR  & Periodic bands  & No  & MRI k-space line dropout \\
\bottomrule
\end{tabular}
\end{table}

\subsection{Results}

As shown in Table~\ref{tab:case_study}, DM-Inpainting, NeuroLIT, PathoPainter, and TPDM emerged as the best-performing methods across the BraTS and TCGA datasets in this case study. Interestingly, two of these methods, NeuroLIT and PathoPainter, are based on LDMs, whereas DM-Inpainting and TPDM rely on DDPMs and score-based diffusion models, respectively. Across all missingness mechanisms, NeuroLIT achieved the highest PSNR and SSIM values on the BraTS dataset. On TCGA, PathoPainter obtained the highest PSNR under the MCAR and MAR mechanisms while also matching the best SSIM scores. Under the MNAR mechanism, however, NeuroLIT outperformed all competing methods in both PSNR and SSIM. Furthermore, NeuroLIT and TPDM were among the fastest approaches evaluated, requiring less than 0.1 seconds per image during inference. These results suggest that latent-space diffusion can provide excellent reconstruction quality while maintaining competitive computational efficiency.

\begin{table}[]
\caption{Overall Peak Signal-to-Noise Ratio (PSNR), Structural Similarity Index Measure (SSIM), and Computational Time results for the BraTS and TCGA datasets. Higher values ($\uparrow$) indicate better performance. The best-performing method for each metric is highlighted in bold, while the second-best result is underlined.}
\label{tab:case_study}
\resizebox{\textwidth}{!}{
\begin{tabular}{llcccccc}
\hline
\multicolumn{8}{c}{\textbf{Missing Completely at Random (MCAR)}}                                                                                 \\ \hline
\multirow{2}{*}{\textbf{\begin{tabular}[c]{@{}l@{}}Diffusion \\ Taxonomy\end{tabular}}} &
  \multicolumn{1}{c}{\multirow{2}{*}{\textbf{\begin{tabular}[c]{@{}c@{}}Inpainting \\ Techniques\end{tabular}}}} &
  \multicolumn{3}{c}{\textbf{BraTS}} &
  \multicolumn{3}{c}{\textbf{TCGA}} \\ \cline{3-8} 
 &
  \multicolumn{1}{c}{} &
  \textbf{PSNR $\uparrow$} &
  \textbf{SSIM $\uparrow$} &
  \multicolumn{1}{l}{\textbf{Time (s)}} &
  \textbf{PSNR $\uparrow$} &
  \textbf{SSIM $\uparrow$} &
  \multicolumn{1}{l}{\textbf{Time (s)}} \\ \hline
DDPM & ArtifactRestoration & 39.630 ± 0.737          & 0.991 ± 0.001          & 0.024 & 33.955 ± 2.553          & 0.980 ± 0.006          & 0.014 \\
DDPM & DM\_Inpainting      & 42.225 ± 1.549          & {\ul 0.996 ± 0.001}    & 1.861 & {\ul 35.774 ± 3.830}    & \textbf{0.986 ± 0.006} & 1.747 \\
DDPM & FastWDM3D           & 31.502 ± 1.183          & 0.913 ± 0.019          & 2.147 & 24.287 ± 1.369          & 0.814 ± 0.104          & 2.147 \\
DDPM & HARP                & 40.926 ± 0.825          & 0.948 ± 0.006          & 2.481 & 34.912 ± 2.752          & 0.981 ± 0.007          & 2.274 \\
SDE  & IR-SDE              & 11.727 ± 0.120          & 0.114 ± 0.022          & 3.714 & 14.014 ± 0.751          & 0.380 ± 0.120          & 3.480 \\
LDM  & NeuroLIT            & \textbf{44.983 ± 1.952} & \textbf{0.998 ± 0.001} & 0.076 & 34.075 ± 3.846          & 0.978 ± 0.008          & 0.060 \\
DDPM & NucleiMix           & 26.670 ± 1.354          & 0.847 ± 0.028          & 4.291 & 20.853 ± 1.700          & 0.692 ± 0.150          & 4.287 \\
LDM  & PathoPainter        & 43.023 ± 1.326          & 0.994 ± 0.002          & 1.863 & \textbf{35.915 ± 3.856} & \textbf{0.986 ± 0.006} & 1.749 \\
SDE  & SDEdit              & 32.455 ± 1.291          & 0.939 ± 0.020          & 1.299 & 25.398 ± 1.703          & 0.835 ± 0.107          & 1.297 \\
SGM  & TPDM                & {\ul 44.259 ± 1.898}    & \textbf{0.998 ± 0.001} & 0.060 & 34.758 ± 3.792          & {\ul 0.982 ± 0.006}    & 0.048 \\ \hline
\multicolumn{8}{c}{\textbf{Missing at Random (MAR)}}                                                                                             \\ \hline
DDPM & ArtifactRestoration & 38.462 ± 0.964          & 0.988 ± 0.002          & 0.023 & 33.699 ± 2.467          & 0.978 ± 0.009          & 0.014 \\
DDPM & DM\_Inpainting      & 42.343 ± 1.362          & {\ul 0.996 ± 0.002}    & 1.861 & {\ul 35.131 ± 0.337}    & \textbf{0.984 ± 0.006} & 1.747 \\
DDPM & FastWDM3D           & 30.938 ± 1.148          & 0.912 ± 0.019          & 2.146 & 23.770 ± 1.434          & 0.813 ± 0.094          & 2.145 \\
DDPM & HARP                & 40.070 ± 0.668          & 0.932 ± 0.005          & 2.482 & 34.400 ± 0.281          & {\ul 0.980 ± 0.008}    & 2.272 \\
SDE  & IR-SDE              & 11.756 ± 0.134          & 0.172 ± 0.022          & 3.710 & 14.024 ± 0.750          & 0.424 ± 0.111          & 3.478 \\
LDM  & NeuroLIT            & \textbf{44.927 ± 1.928} & \textbf{0.998 ± 0.001} & 0.076 & 34.070 ± 3.752          & 0.978 ± 0.008          & 0.060 \\
DDPM & NucleiMix           & 26.363 ± 1.465          & 0.847 ± 0.026          & 4.290 & 20.724 ± 0.171          & 0.709 ± 0.137          & 4.286 \\
LDM  & PathoPainter        & 40.721 ± 0.743          & 0.948 ± 0.010          & 1.861 & \textbf{35.428 ± 0.346} & \textbf{0.984 ± 0.005} & 1.748 \\
SDE  & SDEdit              & 32.401 ± 1.399          & 0.942 ± 0.018          & 1.297 & 24.746 ± 0.189          & 0.834 ± 0.103          & 1.296 \\
SGM  & TPDM                & {\ul 44.565 ± 1.855}    & \textbf{0.998 ± 0.001} & 0.060 & 34.371 ± 3.677          & {\ul 0.980 ± 0.007}    & 0.048 \\ \hline
\multicolumn{8}{c}{\textbf{Missing Not at Random (MNAR)}}                                                                                        \\ \hline
DDPM & ArtifactRestoration & 32.590 ± 0.347          & 0.967 ± 0.016          & 0.013 & 30.469 ± 2.723          & 0.974 ± 0.028          & 0.013 \\
DDPM & DM\_Inpainting      & 20.540 ± 1.693          & 0.913 ± 0.031          & 1.857 & 28.092 ± 3.201          & 0.968 ± 0.035          & 1.745 \\
DDPM & FastWDM3D           & 17.470 ± 0.800          & 0.848 ± 0.017          & 2.150 & 20.870 ± 0.247          & 0.873 ± 0.055          & 2.143 \\
DDPM & HARP                & 24.310 ± 0.254          & 0.942 ± 0.016          & 2.480 & 27.819 ± 3.951          & 0.963 ± 0.037          & 2.272 \\
SDE  & IR-SDE              & 14.227 ± 0.843          & 0.797 ± 0.015          & 3.703 & 14.637 ± 0.712          & 0.625 ± 0.081          & 3.476 \\
LDM &
  NeuroLIT &
  \textbf{39.960 ± 2.321} &
  \textbf{0.986 ± 0.004} &
  0.076 &
  \textbf{35.694 ± 2.651} &
  \textbf{0.991 ± 0.006} &
  0.058 \\
DDPM & NucleiMix           & 17.020 ± 1.062          & 0.805 ± 0.017          & 4.283 & 24.481 ± 4.202          & 0.926 ± 0.062          & 0.428 \\
LDM  & PathoPainter        & 23.700 ± 0.334          & 0.937 ± 0.022          & 1.853 & 30.393 ± 4.051          & 0.961 ± 0.052          & 1.748 \\
SDE  & SDEdit              & 13.966 ± 1.426          & 0.793 ± 0.015          & 1.294 & 26.266 ± 3.184          & 0.924 ± 0.047          & 1.294 \\
SGM  & TPDM                & {\ul 39.877 ± 0.212}    & {\ul 0.985 ± 0.004}    & 0.054 & {\ul 35.279 ± 2.507}    & {\ul 0.990 ± 0.008}    & 0.047 \\ \hline
\end{tabular}
}
\end{table}

Regarding the missingness mechanisms, the relative performance of the leading methods remained stable across MCAR, MAR, and MNAR, despite the different strategies used to generate missing pixels (i.e., masks). In particular, NeuroLIT demonstrated consistently robust performance across all missingness mechanisms, emerging as the strongest overall method in this case study. Under the more challenging MNAR setting, NeuroLIT maintained its superior performance on both the BraTS and TCGA datasets, with TPDM consistently ranking among the top-performing methods.

Overall, DM-Inpainting and TPDM exhibited a particularly interesting pattern, consistently ranking among the best or second-best methods across both datasets despite the substantial differences between the underlying imaging modalities. Specifically, TPDM achieved the highest SSIM on BraTS, whereas DM-Inpainting obtained the highest SSIM on TCGA. For PSNR, NeuroLIT achieved the best performance on BraTS, followed by TPDM, while PathoPainter ranked first on TCGA, with DM-Inpainting placing second. These findings indicate that DM-Inpainting and TPDM provide robust and consistent reconstruction performance across distinct medical imaging domains, highlighting their potential as general-purpose diffusion-based inpainting frameworks.

Although DDPM-based methods represent the majority of the reviewed literature, the strongest performers in our benchmark were predominantly LDM-based approaches. In particular, NeuroLIT achieved the highest overall reconstruction quality on BraTS while maintaining one of the lowest inference times. This finding suggests that latent-space diffusion offers a favorable trade-off between reconstruction fidelity and computational efficiency for medical image inpainting. More broadly, these results demonstrate that the popularity of a diffusion family in the literature does not necessarily translate into superior empirical performance, underscoring the importance of standardized benchmarking across datasets, imaging modalities, and missing data mechanisms.

\section{Conclusions}
\label{sec:conclusions}

This paper presented a systematic review of diffusion-based approaches for medical image inpainting. Following the PRISMA methodology, 60 studies were identified, analyzed, and categorized to provide a comprehensive overview of the current state of the art. To the best of our knowledge, this is one of the first reviews to specifically focus on diffusion models for medical image inpainting while providing both a structured taxonomy and an empirical benchmark of publicly available implementations \cite{kazerouni2023diffusion, azad2026systematic,khosravi2025exploring}.

Based on the reviewed literature, we proposed a taxonomy that organizes diffusion models into their main methodological families, including DDPMs, LDMs, SGMs, and SDE approaches. Our analysis revealed that diffusion models have been successfully applied to a wide range of medical image inpainting tasks, including artifact removal, pseudo-healthy tissue reconstruction, data augmentation, and restoration of missing image regions. Among the identified families, DDPMs and LDMs were the most frequently adopted architectures in the literature.

To complement the systematic review, we conducted a case study benchmarking ten diffusion-based inpainting models on the BraTS and TCGA datasets under a unified experimental protocol. The results demonstrated that diffusion models generally achieve strong reconstruction performance across distinct imaging modalities. Interestingly, although DDPMs constitute the majority of published studies, several of the strongest-performing methods were based on latent diffusion architectures, suggesting that latent-space diffusion may offer a favorable trade-off between reconstruction quality and computational efficiency for medical image inpainting.

Despite the remarkable progress observed in recent years, several challenges remain. Future research directions include improving the interpretability and explainability of diffusion-based reconstructions, developing multi-modal and multi-view inpainting frameworks, investigating the impact of dataset size and diversity on model generalization, and incorporating clinician-centered evaluations to assess the clinical utility of generated reconstructions. Furthermore, the growing emergence of foundation models and large-scale generative architectures presents new opportunities for developing more robust and generalizable medical image inpainting systems.

Overall, the findings of this review demonstrate the growing importance of diffusion models in medical image inpainting and provide researchers with a comprehensive reference for understanding current trends, methodological advances, and future research opportunities in this rapidly evolving field.

\section{Acknowledgments}

This study was financed, in part, by the São Paulo Research Foundation (FAPESP), Brazil, under Grant Nos. 2021/06870-3 and 2024/23791-8. This work was also supported by national funds through FCT -- Fundação para a Ciência e a Tecnologia, I.P., within the scope of the projects UIDB/00326/2025 and UIDP/00326/2025.

Additionally, this research received support from the Portuguese Recovery and Resilience Plan (PRR) through the project C645008882-00000055 -- Center for Responsible AI.

This work is also funded by national funds through FCT – Fundação para a Ciência e a Tecnologia, I.P., under the LASIGE Research Unit, ref. UID/00408/2025,  \url{https://doi.org/10.54499/UID/00408/2025}.

This work was further funded by national funds through FCT -- Foundation for Science and Technology, I.P., under the research unit UID/00326 -- Centre for Informatics and Systems of the University of Coimbra (CISUC), \url{https://doi.org/10.54499/UID/00326/2025}.

\bibliographystyle{ACM-Reference-Format}
\bibliography{acmart}

@article{Vavekanand2026,
  title={A lightweight physics-conditioned diffusion multi-model for medical image reconstruction},
  author={Vavekanand and Raja and Kumar, Ganesh and Kurbanova, Shakhlokhon},
  journal={Biomedical Engineering Communications},
  volume={5},
  number={2},
  pages={12:1--10},
  year={2026},
  publisher={TMR Publishing Group},
  doi={10.53388/bmec2026012}
  }

@InProceedings{Joana2024,
  author={Santos, Joana Cristo and Santos, Miriam Seoane and Abreu, Pedro Henriques},
  title={An Interpretable Human-in-the-Loop Process to Improve Medical Image Classification},
  booktitle={Symposium on Intelligent Data Analysis (IDA)},
  year={2024},
  volume={14641},
  publisher={Springer},
  address={Cham},
  pages={179--190},
  doi={10.1007/978-3-031-58547-0\_15}
}

@article{Elharrouss2024,
  title={Transformer-based image and video inpainting: current challenges and future directions},
  author={Omar Elharrouss and Rafat Damseh and Abdelkader Nasreddine Belkacem and Elarbi Badidi and Abderrahmane Lakas},
  journal={Artificial Intelligence Review},
  year={2024},
  volume={58},
  pages={124:1--45},
  doi={10.1007/s10462-024-11075-9}
}

@article{Santos2025,
  title={The Role of Deep Learning in Medical Image Inpainting: A Systematic Review},
  author={Santos, Joana Cristo and Tom{\'a}s Pereira Alexandre, Hugo and Seoane Santos, Miriam and Henriques Abreu, Pedro},
  journal={ACM Transactions on Computing for Healthcare},
  volume={6},
  number={3},
  pages={1--24},
  year={2025},
  publisher={ACM New York, NY},
  doi={10.1145/3712710}
}

@article{Chen2025,
  author = {Chen, Hang and Xiang, Qian and Hu, Jiaxin and Ye, Meilin and Yu, Chao and Cheng, Hao and Zhang, Lei},
  title = {Comprehensive exploration of diffusion models in image generation: a survey},
  journal = {Artificial Intelligence Review},
  year = {2025},
  volume = {58},
  number = {4},
  pages = {99},
  issn = {1573-7462},
  doi = {10.1007/s10462-025-11110-3},
}

@inbook{SongErmon2019,
author = {Song, Yang and Ermon, Stefano},
title = {Generative modeling by estimating gradients of the data distribution},
year = {2019},
publisher = {Curran Associates Inc.},
address = {Red Hook, NY, USA},
booktitle = {Proceedings of the 33rd International Conference on Neural Information Processing Systems},
articleno = {1067},
pages = {13}
}

@ARTICLE{Croitoru2023,
  author={Croitoru, Florinel-Alin and Hondru, Vlad and Ionescu, Radu Tudor and Shah, Mubarak}, journal={IEEE Transactions on Pattern Analysis \& Machine Intelligence },
  title={Diffusion Models in Vision: A Survey},
  year={2023},
  volume={45},
  number={09},
  pages={10850-10869},
  doi={10.1109/TPAMI.2023.3261988},
  publisher={IEEE Computer Society},
  address={Los Alamitos, CA, USA}
}

@article{Hasan2021,
  author = {Hasan, Md. Kamrul and Alam, Md. Ashraful and Roy, Shidhartho and Dutta, Aishwariya and Jawad, Md Tasnim and Das, Sunanda},
  year = {2021},
  pages = {100799},
  title = {Missing value imputation affects the performance of machine learning: A review and analysis of the literature (2010–2021)},
  volume = {27},
  journal = {Informatics in Medicine Unlocked},
  doi={10.1016/j.imu.2021.100799}
}

@article{azad2026systematic,
  title={{A Systematic Review of Diffusion Models for Medical Image-Based Diagnosis: Methods, Taxonomies, Clinical Integration, Explainability, and Future Directions}},
  author={Azad, Mohammad and Fahad, Nur Mohammad and Raiaan, Mohaimenul Azam Khan and Anik, Tanvir Rahman and Khan, Md Faraz Kabir and Toy{\'e}, Habib Mahamadou K{\'e}l{\'e} and Muhammad, Ghulam},
  journal={Diagnostics},
  volume={16},
  number={2},
  pages={211},
  year={2026},
  publisher={MDPI},
  doi={10.3390/diagnostics16020211}
}

@article{kazerouni2023diffusion,
  title={Diffusion models in medical imaging: A comprehensive survey},
  author={Kazerouni, Amirhossein and Aghdam, Ehsan Khodapanah and Heidari, Moein and Azad, Reza and Fayyaz, Mohsen and Hacihaliloglu, Ilker and Merhof, Dorit},
  journal={Medical image analysis},
  volume={88},
  pages={102846},
  year={2023},
  publisher={Elsevier},
  doi={10.1016/j.media.2023.102846}
}

@article{khosravi2025exploring,
  title={Exploring the potential of generative artificial intelligence in medical image synthesis: opportunities, challenges, and future directions},
  author={Khosravi, Bardia and Purkayastha, Saptarshi and Erickson, Bradley J and Trivedi, Hari M and Gichoya, Judy W},
  journal={The Lancet Digital Health},
  year={2025},
  publisher={Elsevier},
  volume={7},
  number={9},
  pages={100890},
  doi={10.1016/j.landig.2025.100890}
}

@InProceedings{Bae2024,
  author = {Bae, Juyoung and Tong, Elizabeth and Chen, Hao},
  title = {Conditional Diffusion Model for Versatile Temporal Inpainting in 4D Cerebral CT Perfusion Imaging},
  booktitle = {International Conference on Medical Image Computing and Computer-Assisted Intervention (MICCAI)},
  year = {2024},
  publisher = {Springer},
  address={Cham},
  volume = {15002},
  pages = {67 -- 77},
  doi={10.1007/978-3-031-72069-7\_7}
}

@article{Dong2024,
  title = {ClarityDiffuseNet: Enhancing fundus image quality under black shadows with diffusion model-based research},
  journal = {Pattern Recognition Letters},
  volume = {186},
  pages = {279-285},
  year = {2024},
  doi = {10.1016/j.patrec.2024.10.012},
  author = {Jiadi Dong and Tianwei Qian and Yuxian Jiang and Lei Bi and Jinman Kim and Lisheng Wang and Xun Xu},
}

@InProceedings{Durrer2025,
  author={Durrer, Alicia and Wolleb, Julia and Bieder, Florentin and Friedrich, Paul and Melie-Garcia, Lester and Ocampo Pineda, Mario Alberto and Bercea, Cosmin I. and Hamamci, Ibrahim Ethem and Wiestler, Benedikt and Piraud, Marie and Yaldizli, Oezguer and Granziera, Cristina and Menze, Bjoern and Cattin, Philippe C. and Kofler, Florian},
  title={Denoising Diffusion Models for 3D Healthy Brain Tissue Inpainting},
  booktitle={Deep Generative Models (DGM4MICCAI)},
  year={2025},
  publisher={Springer},
  address={Cham},
  pages={87--97},
  doi={10.1007/978-3-031-72744-3\_9}
}

@inproceedings{Durrer2023,
  author = {Durrer, Alicia and Cattin, Philippe C. and Wolleb, Julia},
  title = {Denoising Diffusion Models for Inpainting of Healthy Brain Tissue},
  year = {2023},
  publisher = {Springer},
  address = {Cham},
  doi = {10.1007/978-3-031-76163-8_4},
  booktitle = {Brain Tumor Segmentation, and Cross-Modality Domain Adaptation for Medical Image Segmentation (crossMoDA BraTS)},
  pages = {35–45},
  numpages = {11}
}

@InProceedings{Durrer2026,
  author={Durrer, Alicia and Bieder, Florentin and Friedrich, Paul and Menze, Bjoern and Cattin, Philippe C. and Kofler, Florian},
  title={fastWDM3D: Fast and Accurate 3D Healthy Tissue Inpainting},
  booktitle={Deep Generative Models (DGM4MICCAI)},
  year={2026},
  publisher={Springer},
  address={Cham},
  pages={171--181},
  doi={10.1007/978-3-032-05472-2\_17}
  }

@INPROCEEDINGS{Kebaili2025,
  author={Kebaili, Aghiles and Lapuyade-Lahorgue, Jérôme and Vera, Pierre and Ruan, Su},
  booktitle={International Symposium on Biomedical Imaging (ISBI)}, 
  title={AMM-Diff: Adaptive Multi-Modality Diffusion Network for Missing Modality Imputation}, 
  year={2025},
  volume={},
  number={},
  pages={1-4},
  publisher={IEEE},
  address={Houston, TX, USA},
  doi={10.1109/ISBI60581.2025.10980985}
  }

@InProceedings{Kwark2026,
  author={Kwark, Dou Hoon and Luo, Shirui and Zhu, Xiyue and Li, Yudu and Liang, Zhi-Pei and Kindratenko, Volodymyr},
  title={Hierarchical Diffusion Framework for Pseudo-healthy Brain MRI Inpainting with Enhanced 3D Consistency},
  booktitle={Deep Generative Models (DGM4MICCAI)},
  year={2026},
  publisher={Springer},
  address={Cham},
  pages={78--88},
  doi={10.1007/978-3-032-05472-2\_8}
}

@article{Kim2026,
  title = {Visual prompt tuning for task-flexible medical image synthesis},
  journal = {Computer Methods and Programs in Biomedicine},
  volume = {277},
  pages = {109244},
  year = {2026},
  doi = {10.1016/j.cmpb.2026.109244},
  author = {Jonghun Kim and Hyunjin Park}
}

@InProceedings{Fuchs2024,
  title = {HARP: Unsupervised Histopathology Artifact Restoration},
  author = {Fuchs, Moritz and Sivakumar, Ssharvien Kumar R and Sch\"ober, Mirko and Woltering, Niklas and Eich, Marie-Lisa and Schweizer, Leonille and Mukhopadhyay, Anirban},
  booktitle = {International Conference on Medical Imaging with Deep Learning (MIDL)},
  pages = {465--479},
  year = {2024},
  volume = {250},
  publisher = {Proceedings of Machine Learning Research},
  address={Paris, France},
  url={https://proceedings.mlr.press/v250/fuchs24a.html}
}

@INPROCEEDINGS{Kadi2023,
  author={Kadi, Hocine and Sourget, Théo and Kawczynski, Marzena and Bendjama, Sara and Grollemund, Bruno and Bloch-Zupan, Agnés},
  booktitle={International Conference on Computational Science and Computational Intelligence (CSCI)}, 
  title={Segmentation, and Numbering in Oral Rare Diseases: Focus on Data Augmentation and Inpainting Techniques}, 
  year={2023},
  volume={},
  number={},
  pages={1358-1363},
  publisher={IEEE},
  address= {Las Vegas, NV, USA},
  doi={10.1109/CSCI62032.2023.00298}}

@INPROCEEDINGS{Jiang2025,
  author={Jiang, Yankai and Zhang, Peng and Yang, Donglin and Tian, Yuan and Lin, Hai and Wang, Xiaosong},
  booktitle={Conference on Computer Vision and Pattern Recognition (CVPR)}, 
  title={Advancing Generalizable Tumor Segmentation with Anomaly-Aware Open-Vocabulary Attention Maps and Frozen Foundation Diffusion Models}, 
  year={2025},
  publisher={IEEE/CVF},
  address={Nashville, TN, USA},
  pages={25971-25981},
  doi={10.1109/CVPR52734.2025.02419}
}

@INPROCEEDINGS{GuoTao2024,
  author={Guo, Lianghu and Tao, Tianli and Cai, Xinyi and Zhu, Zihao and Huang, Jiawei and Zhu, Lixuan and Gu, Zhuoyang and Tang, Haifeng and Zhou, Rui and Han, Siyan and Liang, Yan and Yang, Qing and Shen, Dinggang and Zhang, Han},
  booktitle={International Symposium on Biomedical Imaging (ISBI)}, 
  title={Cas-DiffCom: Cascaded Diffusion Model for Infant Longitudinal Super-Resolution 3D Medical Image Completion}, 
  year={2024},
  pages={1-4},
  doi={10.1109/ISBI56570.2024.10635663},
  publisher={IEEE},
  address={Athens, Greece}
  }

@INPROCEEDINGS{Guo2025,
  author={Guo, Jiaqi and López-Tapia, Santiago and Katsaggelos, Aggelos K.},
  booktitle={International Conference on Image Processing (ICIP)}, 
  title={Advancing Limited-Angle CT Reconstruction through Diffusion-Based Sinogram Completion}, 
  year={2025},
  volume={},
  number={},
  pages={2211-2216},
  doi={10.1109/ICIP55913.2025.11084640},
  address={Anchorage, AK, USA},
  publisher={IEEE}
  }

@Article{Koo2023,
  AUTHOR = {Koo, JaKeoung},
  TITLE = {Sinogram Upsampling via Sub-Riemannian Diffusion with Adaptive Weighting},
  JOURNAL = {Electronics},
  VOLUME = {12},
  YEAR = {2023},
  NUMBER = {21},
  pages = {4503:1--15},
  DOI = {10.3390/electronics12214503}
}

@INPROCEEDINGS{Ji2023,
  author={Ji, Bangning and He, Gang},
  booktitle={International Conference on Bioinformatics and Biomedicine (BIBM)}, 
  title={A Novel Diffusion-Model-Based Bone Scan Image Inpainting Algorithm}, 
  year={2023},
  address={Istanbul, Turkiye},
  publisher={IEEE},
  pages={4907-4909},
  doi={10.1109/BIBM58861.2023.10386039}
  }

@INPROCEEDINGS{Kropp2024,
  author={Kropp, Jan-Oliver and Schiffer, Christian and Amunts, Katrin and Dickscheid, Timo},
  booktitle={International Symposium on Biomedical Imaging (ISBI)}, 
  title={Denoising Diffusion Probabilistic Models for Image Inpainting of Cell Distributions in The Human Brain}, 
  year={2024},
  volume={},
  number={},
  pages={1-5},
  publisher={IEEE},
  address={Athens, Greece},
  doi={10.1109/ISBI56570.2024.10635384}}

@ARTICLE{Karageorgos2024,
  author={Karageorgos, Grigorios M. and Zhang, Jiayong and Peters, Nils and Xia, Wenjun and Niu, Chuang and Paganetti, Harald and Wang, Ge and De Man, Bruno},
  journal={IEEE Transactions on Medical Imaging}, 
  title={A Denoising Diffusion Probabilistic Model for Metal Artifact Reduction in CT}, 
  year={2024},
  volume={43},
  number={10},
  pages={3521-3532},
  doi={10.1109/TMI.2024.3416398}}

@INPROCEEDINGS{Li2025,
  author={Li, Xingnan and Rana, Priyanka and Gide, Tuba N and Adegoke, Nurudeen A and Mao, Yizhe and Wilmott, James S and Liu, Sidong},
  booktitle={International Conference on Bioinformatics and Biomedicine (BIBM)}, 
  title={OS2CR-Diff: A Self-Refining Diffusion Framework for CD8 Imputation from One-Step Inference to Conditional Representation}, 
  year={2025},
  volume={},
  number={},
  pages={1062-1069},
  publisher={IEEE},
  address={Wuhan, China},
  doi={10.1109/BIBM66473.2025.11356570}}

@Article{Hung2023,
  AUTHOR = {Hung, Alex Ling Yu and Zhao, Kai and Zheng, Haoxin and Yan, Ran and Raman, Steven S. and Terzopoulos, Demetri and Sung, Kyunghyun},
  TITLE = {Med-cDiff: Conditional Medical Image Generation with Diffusion Models},
  JOURNAL = {Bioengineering},
  VOLUME = {10},
  YEAR = {2023},
  NUMBER = {11},
  pages = {1258:1--13},
  PubMedID = {38002382},
  DOI = {10.3390/bioengineering10111258}
}

@InProceedings{Ji2024,
  author={Ji, Bangning and He, Gang and Chen, Zhengguo and Zhao, Ling},
  title={A Novel Diffusion-Model-Based OCT Image Inpainting Algorithm for Wide Saturation Artifacts},
  booktitle={Pattern Recognition and Computer Vision (PRCV)},
  year={2024},
  publisher={Springer},
  address={Singapore},
  pages={284--295},
  doi={10.1007/978-981-99-8558-6\_24}
  }

@INPROCEEDINGS{Kalantar2023,
  author={Kalantar, Reza and Lin, Gigin and Winfield, Jessica M and Messiou, Christina and Koh, Dow-Mu and Blackledge, Matthew D},
  booktitle={International Conference on Medical Artificial Intelligence (MedAI)}, 
  title={MED-INPAINT: Medical Image Synthesis Using Multi-Level Conditional Inpainting with a Denoising Diffusion Probabilistic Model and Adaptive Contrast Priors}, 
  year={2023},
  publisher = {IEEE Computer Society},
  address = {Los Alamitos, CA, USA},
  pages={403-413},
  doi={10.1109/MedAI59581.2023.00061}
  }

@InProceedings{Lei2025,
    author = {Lei, Wenhui AND Tian, Hengrui AND Dai, Linrui AND Chen, Hanyu AND Zhang, Xiaofan},
    title = {LesionDiffusion: Towards Text-controlled General Lesion Synthesis},
    booktitle = {International Conference on Medical Image Computing and Computer Assisted Intervention (MICCAI)},
    year = {2025},
    publisher = {Springer},
    volume = {15964},
    address={Cham},
    pages = {327 -- 336},
    doi={10.1007/978-3-032-04971-1\_31}
}

@InProceedings{Graf24a,
  title = {Modeling the acquisition shift between axial and sagittal MRI for diffusion superresolution to enable axial spine segmentation},
  author = {Graf, Robert and M\"oller, Hendrik and McGinnis, Julian and R\"uhling, Sebastian and Weihrauch, Maren and Atad, Matan and Shit, Suprosanna and Menze, Bjoern and M\"uhlau, Mark and Paetzold, Johannes C. and Rueckert, Daniel and Kirschke, Jan},
  booktitle = {International Conference on Medical Imaging with Deep Learning (MIDL)},
  pages = {520--537},
  year = {2024},
  volume = {250},
  publisher = {Proceedings of Machine Learning Research},
  address={Paris, France},
  url={https://proceedings.mlr.press/v250/graf24a.html}
}

@InProceedings{He2026,
  author={He, Shidan and Hu, Enyuan and Tang, Zixuan and Chen, Bin and Yu, Dongdong and Hong, Yuan and Liu, Zhenzhong and Li, Mengtang and Liu, Lei and Zhao, Shen},
  title={MedSoft-Diffusion: Medical Semantic-Guided Diffusion Model with Soft Mask Conditioning for Vertebral Disease Diagnosis},
  booktitle={International Conference on Medical Image Computing and Computer Assisted Intervention (MICCAI)},
  year={2025},
  publisher={Springer},
  address={Cham},
  pages={333--342},
  doi={10.1007/978-3-032-05182-0\_33}
  }

@InProceedings{Jeong2023,
  author={Jeong, Jiheon and Kim, Ki Duk and Nam, Yujin and Cho, Kyungjin and Kang, Jiseon and Hong, Gil-Sun and Kim, Namkug},
  title={Generating High-Resolution 3D CT with 12-Bit Depth Using a Diffusion Model with Adjacent Slice and Intensity Calibration Network},
  booktitle={International Conference on Medical Image Computing and Computer Assisted Intervention (MICCAI)},
  year={2023},
  publisher={Springer},
  address={Cham},
  pages={366--375},
  doi={10.1007/978-3-031-43999-5\_35}
}

@inproceedings{Liu2024,
  title={3D skull completion via two-stage conditional diffusion-based signed distance fields},
  author={Liu, Zhenhong and Ru, Xudong and Wang, Xingce and Wu, Zhongke and Zhu, Yi-Cheng and Zhang, Chong and Frangi, Alejandro F},
  booktitle={IEEE International Conference on Bioinformatics and Biomedicine (BIBM)},
  pages={2204--2209},
  year={2024},
  publisher={IEEE},
  address={Lisbon, Portugal},
  doi={10.1109/BIBM62325.2024.10822061}
}

@inproceedings{Liu2026,
  title={Pathopainter: Augmenting histopathology segmentation via tumor-aware inpainting},
  author={Liu, Hong and Yang, Haosen and Huijben, Evi MC and Schuiveling, Mark and Su, Ruisheng and Pluim, Josien PW and Veta, Mitko},
  booktitle={International Conference on Medical Image Computing and Computer-Assisted Intervention (MICCAI)},
  pages={408--417},
  year={2025},
  publisher={Springer},
  address={Cham},
  doi={10.1007/978-3-032-05325-1\_39}
}

@inproceedings{Ma2024,
  title={Generalize Polyp Segmentation Via Inpainting Across Diverse Backgrounds and Pseudo-Mask Refinement},
  author={Ma, Jiajian and Lu, Fangqi and Huang, Silin and Wu, Song and Li, Zhen},
  booktitle={International Symposium on Biomedical Imaging (ISBI)},
  pages={1--5},
  year={2024},
  doi={10.1109/ISBI56570.2024.10635889},
  publisher={IEEE},
  address={Athens, Greece}
}

@inproceedings{Mei2023,
  title={Metal inpainting in CBCT projections using score-based generative model},
  author={Mei, Siyuan and Fan, Fuxin and Maier, Andreas},
  booktitle={International Symposium on Biomedical Imaging (ISBI)},
  pages={1--5},
  year={2023},
  publisher={IEEE},
  address={Cartagena, Colombia},
  doi={10.1109/ISBI53787.2023.10230638}
}

@article{Montoya2024,
  title={MAM-E: Mammographic synthetic image generation with diffusion models},
  author={Montoya-del-Angel, Ricardo and Sam-Millan, Karla and Vilanova, Joan C and Mart{\'\i}, Robert},
  journal={Sensors},
  volume={24},
  number={7},
  pages={2076},
  year={2024},
  publisher={MDPI},
  doi={10.3390/s24072076}
}

@inproceedings{Montoya2025,
  title={ELK: Enhanced Learning Through Cross-Modal Knowledge Transfer for Lesion Detection in Limited-Sample Contrast-Enhanced Mammography Datasets},
  author={Montoya-del-Angel, Ricardo and Elbatel, Marawan and Castillo-Lopez, Jorge Patricio and Villase{\~n}or-Navarro, Yolanda and Brandan, Maria-Ester and Marti, Robert},
  booktitle={Deep Breast Workshop on AI and Imaging for Diagnostic and Treatment Challenges in Breast Care},
  pages={221--231},
  year={2024},
  publisher={Springer},
  address={Cham},
  doi={10.1007/978-3-031-77789-9\_22}
}

@inproceedings{Olsen2025,
  title={Unsupervised detection of fetal brain anomalies using denoising diffusion models},
  author={Olsen, Markus Ditlev Sj{\o}gren and Ambsdorf, Jakob and Lin, Manxi and Taks{\o}e-Vester, Caroline and Svendsen, Morten Bo S{\o}ndergaard and Christensen, Anders Nymark and Nielsen, Mads and Tolsgaard, Martin Gr{\o}nneb{\ae}k and Feragen, Aasa and Pegios, Paraskevas},
  booktitle={International Workshop on Advances in Simplifying Medical Ultrasound},
  pages={209--219},
  year={2024},
  publisher={Springer},
  address={Cham},
  doi={10.1007/978-3-031-73647-6_20}
}

@article{Pollak2025,
  title={FastSurfer-LIT: Lesion inpainting tool for whole-brain MRI segmentation with tumors, cavities, and abnormalities},
  author={Pollak, Clemens and K{\"u}gler, David and Bauer, Tobias and R{\"u}ber, Theodor and Reuter, Martin},
  journal={Imaging Neuroscience},
  volume={3},
  pages={imag\_a\_00446: 1--25},
  year={2025},
  publisher={MIT Press},
  address={Cambridge, Massachusetts, USA},
  doi={10.1162/imag_a_00446}
}

@article{Prochazka2026,
  title={Domain adaptation of stable diffusion for ultrasound inpainting: a synthetic data approach for enhanced thyroid nodule segmentation},
  author={Prochazka, Antonin and Zeman, Jan},
  journal={Journal of Biomedical Informatics},
  volume={173},
  number={104963},
  pages={1--9},
  year={2025},
  publisher={Elsevier},
  doi={10.1016/j.jbi.2025.104963}
}

@article{Qiao2025,
  title={ConNeCT: weakly supervised corneal confocal microscopy image inpainting network based on a diffusion model},
  author={Qiao, Qincheng and Hou, Xinguo},
  journal={Biomedical Optics Express},
  volume={16},
  number={7},
  pages={2615--2630},
  year={2025},
  publisher={Optica Publishing Group},
  doi={10.1364/BOE.562924}
}

@inproceedings{Schaub2026,
  title={3D CBCT Artefact Removal Using Perpendicular Score-Based Diffusion Models},
  author={Schaub, Susanne and Bieder, Florentin and Oliveira, Matheus L and Wang, Yulan and Dagassan-Berndt, Dorothea and Bornstein, Michael M and Cattin, Philippe C},
  booktitle={Deep Generative Models (DGM4MICCAI)},
  pages={244--253},
  year={2025},
  publisher={Springer},
  address={Cham},
  doi={10.1007/978-3-032-05472-2_24}
}

@article{Shi2025,
  title={Generative Inpainting-Based Anomaly Detection for CT Liver Tumor Detection},
  author={Shi, Yongyi and Niu, Chuang and Simpson, Amber L and De Man, Bruno and Do, Richard and Wang, Ge},
  journal={IEEE Transactions on Radiation and Plasma Medical Sciences},
  year={2025},
  publisher={IEEE},
  volume={9},
  number={8},
  pages={1051--1061},
  doi={10.1109/TRPMS.2025.3551946}
}

@article{Tamang2025,
  title={Development of Preprocessing Stage for Early Cervical Cancer Detection Using UNET Diffusion Model},
  author={Tamang, Parimala and Thatal, Annet and Gupta11, Mousumi and Bhattacharjee, Snehashish},
  journal={Journal of Image and Graphics},
  volume={13},
  number={3},
  year={2025},
  pages={245--252},
  doi={10.18178/joig.13.3.245-252}
}

@inproceedings{Tao2025,
  title={DiffKAN-Inpainting: KAN-based diffusion model for brain tumor inpainting},
  author={Tao, Tianli and Wang, Ziyang and Zhang, Han and Arvanitis, Theodoros N and Zhang, Le},
  booktitle={IEEE International Symposium on Biomedical Imaging (ISBI)},
  pages={1--4},
  year={2025},
  publisher={IEEE},
  address={ Houston, TX, USA},
  doi={10.1109/ISBI60581.2025.10980990}
}

@inproceedings{Tong2023,
  title={Data-consistent unsupervised diffusion model for metal artifact reduction},
  author={Tong, Zhan and Wu, Zhan and Yang, Yang and Mao, Weilong and Wang, Shijie and Li, Yinsheng and Chen, Yang},
  booktitle={IEEE International Conference on Bioinformatics and Biomedicine (BIBM)},
  pages={1467--1472},
  year={2023},
  publisher={IEEE},
  address={Istanbul, Turkiye},
  doi={10.1109/BIBM58861.2023.10385300}
}

@inproceedings{Wang2024_1,
  title={Histology Image Artifact Restoration with Lightweight Transformer Based Diffusion Model},
  author={Wang, Chong and He, Zhenqi and He, Junjun and Ye, Jin and Shen, Yiqing},
  booktitle={International Conference on Artificial Intelligence in Medicine (AIME)},
  pages={81--89},
  year={2024},
  publisher={Springer},
  address={Cham},
  doi={10.1007/978-3-031-66535-6\_9}
}

@article{Wang2024,
  title={DSIS-DPR: Structured instance segmentation and diffusion prior refinement for dental anatomy learning},
  author={Wang, Xianyun and Wang, Linhong and Yang, Zhenchen and Zhou, Jiacong and Zheng, Yuchen and Chen, Feng and Hong, Richang and Yu, Jun and Yang, Fan},
  journal={IEEE Transactions on Multimedia},
  volume={26},
  pages={9464--9476},
  year={2024},
  publisher={IEEE},
  doi={10.1109/TMM.2024.3394777}
}

@article{Wang2025,
  title={NucleiMix: Realistic data augmentation for nuclei instance segmentation},
  author={Wang, Jiamu and Kwak, Jin Tae},
  journal={Computers in Biology and Medicine},
  volume={196},
  pages={110901},
  year={2025},
  publisher={Elsevier},
  doi={10.1016/j.compbiomed.2025.110901}
}

@article{Wehrli2025,
  title={Generating 3D pseudo-healthy knee MR images to support trochleoplasty planning},
  author={Wehrli, Michael and Durrer, Alicia and Friedrich, Paul and Buchakchiyskiy, Volodimir and Mumme, Marcus and Li, Edwin and Lehoczky, Gyozo and Hasler, Carol C and Cattin, Philippe C},
  journal={International Journal of Computer Assisted Radiology and Surgery},
  volume={20},
  number={6},
  pages={1059--1066},
  year={2025},
  publisher={Springer},
  address={Cham},
  doi={10.1007/s11548-025-03343-y}
}

@article{Wu2025,
  title={UPGRADE-Net: Unsupervised Sinogram-domain Data-Consistent Network for Metal Artifact Reduction},
  author={Wu, Zhan and Zhang, Yikun and Guo, Yongjie and Tang, Hui and Li, Yinsheng and Shu, Huazhong and Xi, Yan and Zhang, Yi and Coatrieux, Gouenou and Chen, Yang},
  journal={IEEE Transactions on Medical Imaging},
  year={2025},
  volume={},
  number={},
  pages={1--12},
  publisher={IEEE},
  doi={10.1109/TMI.2025.3630832}
}

@article{WuZhong2025,
  title={PRAISE-Net: Deep projection-domain data-consistent learning network for CBCT metal artifact reduction},
  author={Wu, Zhan and Zhong, Xinyun and Lyu, Tianling and Xi, Yan and Ji, Xu and Zhang, Yi and Xie, Shipeng and Yu, Hengyong and Chen, Yang},
  journal={IEEE Transactions on Instrumentation and Measurement},
  year={2025},
  volume={74},
  number={},
  pages={1-13},
  publisher={IEEE},
  doi={10.1109/TIM.2025.3551446}
}

@article{Xu2026,
  title={B2E-CDG: Conditional diffusion-based for label-free OCT angiography artifact removal and robust vascular reconstruction},
  author={Xu, Jing and Fu, Suzhong and Xing, Jiwei and Xue, Linyan and Zhao, Qingliang},
  journal={Artificial Intelligence in Medicine},
  volume={173},
  pages={103345:1--10},
  year={2025},
  publisher={Elsevier},
  doi={10.1016/j.artmed.2025.103345}
}

@inproceedings{Yang2025,
  title={MIA: Masked Inpainting-Based Image Augmentation with Diffusion Models for Enhanced Dermatology Image Classification},
  author={Yang, Guohao and Gong, Yanmin and Guo, Yuanxiong},
  booktitle={International Conference on Connected Health: Applications, Systems and Engineering Technologies (CHASE)},
  publisher = {ACM/IEEE},
  address = {New York, NY, USA},
  pages={441--446},
  year={2025},
  doi={10.1145/3721201.3725430}
}

@inproceedings{Zhang2024_1,
  title={Bi-directional MS lesion filling and synthesis using denoising diffusion implicit model-based lesion repainting},
  author={Zhang, Jinwei and Zuo, Lianrui and Liu, Yihao and Remedios, Samuel and Landman, Bennett A and Prince, Jerry L and Carass, Aaron},
  booktitle={Medical Imaging 2025: Image Processing},
  volume={13406},
  pages={217--223},
  year={2025},
  publisher={SPIE},
  address={Bellingham, WA, USA},
  doi={10.1117/12.3047407}
}

@inproceedings{Zhang2025,
  title={LeFusion: Controllable Pathology Synthesis via Lesion-Focused Diffusion Models},
  author={Zhang, Hantao and Liu, Yuhe and Yang, Jiancheng and Wan, Shouhong and Wang, Xinyuan and Peng, Wei and Fua, Pascal},
  booktitle={International Conference on Learning Representations (ICLR)},
  year={2025},
  pages={1--22},
  address={Singapore},
  publisher={OpenReview},
  url={https://openreview.net/forum?id=3b9SKkRAKw}
}

@inproceedings{Zhu2024,
  title={LoCI-DiffCom: Longitudinal Consistency-Informed Diffusion Model for 3D Infant Brain Image Completion},
  author={Zhu, Zihao and Tao, Tianli and Tao, Yitian and Deng, Haowen and Cai, Xinyi and Wu, Gaofeng and Wang, Kaidong and Tang, Haifeng and Zhu, Lixuan and Gu, Zhuoyang and others},
  booktitle={International Conference on Medical Image Computing and Computer-Assisted Intervention (MICCAI)},
  pages={249--258},
  year={2024},
  publisher={Springer},
  address={Cham},
  doi={10.1007/978-3-031-72069-7_24}
}

@inproceedings{Zuo2025,
  title={Robust body composition analysis by generating 3D CT volumes from limited 2D slices},
  author={Zuo, Lianrui and Yu, Xin and Su, Dingjie and Xu, Kaiwen and Krishnan, Aravind R and Liu, Yihao and Bao, Shunxing and Maldonado, Fabien and Ferrucci, Luigi and Landman, Bennett A},
  booktitle={Medical Imaging 2025: Image Processing},
  volume={13406},
  pages={101--108},
  year={2025},
  publisher={SPIE},
  address={Bellingham, WA, USA},
  doi={10.1117/12.3047039}
}

@inproceedings{Ho2020,
author = {Ho, Jonathan and Jain, Ajay and Abbeel, Pieter},
title = {Denoising diffusion probabilistic models},
year = {2020},
isbn = {9781713829546},
publisher = {Curran Associates Inc.},
address = {Red Hook, NY, USA},
booktitle = {Proceedings of the 34th International Conference on Neural Information Processing Systems},
articleno = {574},
numpages = {12},
location = {Vancouver, BC, Canada},
series = {NIPS '20}
}

@INPROCEEDINGS {Rombach2022,
author = { Rombach, Robin and Blattmann, Andreas and Lorenz, Dominik and Esser, Patrick and Ommer, Bjorn },
booktitle = {Conference on Computer Vision and Pattern Recognition (CVPR) },
title = {High-Resolution Image Synthesis with Latent Diffusion Models},
year = {2022},
pages = {10674-10685},
doi = {10.1109/CVPR52688.2022.01042},
publisher = {IEEE/CVF},
address = {Los Alamitos, CA, USA}
}

@article{ANDERSON1982,
title = {Reverse-time diffusion equation models},
journal = {Stochastic Processes and their Applications},
volume = {12},
number = {3},
pages = {313-326},
year = {1982},
issn = {0304-4149},
doi = {https://doi.org/10.1016/0304-4149(82)90051-5},
author = {Brian D.O. Anderson},
}

@inproceedings{song2019sliced,
  author    = {Yang Song and
               Sahaj Garg and
               Jiaxin Shi and
               Stefano Ermon},
  title     = {{Sliced Score Matching: A Scalable Approach to Density and Score Estimation}},
  booktitle = {Conference on Uncertainty in Artificial Intelligence (UAI)},
  pages     = {204},
  year      = {2019},
  publisher = {PMLR},
  address = {Tel Aviv, Israel}
}

@article{Feuerriegel2023,
author = {Feuerriegel, Stefan and Hartmann, Jochen and Janiesch, Christian and Zschech, Patrick},
year = {2024},
month = {02},
volume = {66},
pages = {111--126},
issn      = {1867-0202},
title = {Generative AI},
journal = {Business \& Information Systems Engineering},
doi = {10.1007/s12599-023-00834-7}
}

@article{sengar2025generative,
  author = {Sengar, Sandeep Singh and Hasan, Affan Bin and Kumar, Sanjay and Carroll, Fiona},
  title = {Generative artificial intelligence: a systematic review and applications},
  journal = {Multimedia Tools and Applications},
  volume = {84},
  pages = {23661--23700},
  year = {2025},
  doi = {10.1007/s11042-024-20016-1},
  url = {https://doi.org/10.1007/s11042-024-20016-1}
}

@article{Banh2023,
  author  = {Banh, L. and Strobel, G.},
  title   = {Generative artificial intelligence},
  journal = {Electronic Markets},
  year    = {2023},
  volume  = {33},
  number  = {1},
  pages   = {63},
  doi     = {10.1007/s12525-023-00680-1},
  url     = {https://doi.org/10.1007/s12525-023-00680-1},
  month   = {Dec}
}

@article{liu2024lesion,
  title={Lesion region inpainting: an approach for pseudo-healthy image synthesis in intracranial infection imaging},
  author={Liu, Xiaojuan and Xiang, Cong and Lan, Libin and Li, Chuan and Xiao, Hanguang and Liu, Zhi},
  journal={Frontiers in Microbiology},
  volume={15},
  pages={1453870},
  year={2024},
  doi={10.3389/fmicb.2024.1453870}
}

@article{quan2024deep,
  title={Deep learning-based image and video inpainting: A survey},
  author={Quan, Weize and Chen, Jiaxi and Liu, Yanli and Yan, Dong-Ming and Wonka, Peter},
  journal={International Journal of Computer Vision},
  volume={132},
  number={7},
  pages={2367--2400},
  year={2024},
  publisher={Springer},
  doi={10.1007/s11263-023-01977-6}
}

@ARTICLE{10521640,
  author={Bengesi, Staphord and El-Sayed, Hoda and Sarker, MD Kamruzzaman and Houkpati, Yao and Irungu, John and Oladunni, Timothy},
  journal={IEEE Access}, 
  title={{Advancements in Generative AI: A Comprehensive Review of GANs, GPT, Autoencoders, Diffusion Model, and Transformers}}, 
  year={2024},
  volume={12},
  number={},
  pages={69812-69837},
  doi={10.1109/ACCESS.2024.3397775}
  }

@inproceedings{ xiao2022tackling,
    title={Tackling the Generative Learning Trilemma with Denoising Diffusion {GAN}s},
    author={Zhisheng Xiao and Karsten Kreis and Arash Vahdat},
    booktitle={International Conference on Learning Representations (ICLR)},
    year={2022},
    url={https://openreview.net/forum?id=JprM0p-q0Co},
    pages={1--28},
    publisher={OpenReview}
}

@article{ghosh2024class,
  title={The class imbalance problem in deep learning},
  author={Ghosh, Kushankur and Bellinger, Colin and Corizzo, Roberto and Branco, Paula and Krawczyk, Bartosz and Japkowicz, Nathalie},
  journal={Machine Learning},
  volume={113},
  number={7},
  pages={4845--4901},
  year={2024},
  publisher={Springer},
  doi={10.1007/s10994-022-06268-8}
}

@inproceedings{ferreira2025achieving,
  title={Achieving Over 10$\times$ Faster Sample Generation with Conditional Denoising Diffusion},
  author={Ferreira, Andr{\'e} and Luijten, Gijs and Hinrichs-Puladi, Behrus and Kleesiek, Jens and Alves, Victor and Egger, Jan},
  booktitle={International Conference on Medical Image Computing and Computer-Assisted Intervention (MICCAI)},
  pages={123--132},
  year={2025},
  publisher={Springer},
  address={Cham},
  doi={10.1007/978-3-032-16370-7\_11}
}

@inproceedings{hahne2026ai,
  title={AI-Driven Prognosis Visualization and Behavioral Change Induction for Diabetic Foot Ulcers via XR Glasses},
  author={Hahne, Jason and Chen, Ma and Kanzhi, Wu and Wei, Li},
  booktitle={International Conference on Artificial Intelligence and eXtended and Virtual Reality (AIxVR)},
  pages={193--197},
  year={2026},
  publisher={IEEE},
  address={Osaka, Japan},
  doi={10.1109/AIxVR67263.2026.00033}
}

@inproceedings{dai2025context,
  title={Context-Aware Healthy Brain Inpainting: A Multi-stage DDIM Approach for the BraTS 2025 Challenge},
  author={Dai, Yaxuan and Bi, Yuan and Navab, Nassir and Jiang, Zhongliang},
  booktitle={International Conference on Medical Image Computing and Computer-Assisted Intervention (MICCAI)},
  pages={158--167},
  year={2025},
  publisher={Springer},
  address={Cham},
  doi={10.1007/978-3-032-16370-7\_14}
}

@inproceedings{nazir2025diffusion,
  title={Diffusion-Based Data Augmentation for Medical Image Segmentation},
  author={Nazir, Maham and Aqeel, Muhammad and Setti, Francesco},
  booktitle={International Conference on Computer Vision (ICCV)},
  pages={1330--1339},
  year={2025},
  publisher={IEEE/CVF},
  address={Honolulu, HI, USA},
  doi={10.1109/ICCVW69036.2025.00143}
}

@inproceedings{koch2025local,
  title={Local Brain Tumour Inpainting Using Diffusion Transformers},
  author={Koch, Alexander and Aydin, Orhun Utku and Hilbert, Adam and Frey, Dietmar},
  booktitle={International Conference on Medical Image Computing and Computer-Assisted Intervention (MICCAI)},
  pages={133--145},
  year={2025},
  publisher={Springer},
  address={Cham},
  doi={10.1007/978-3-032-16370-7\_12}
}

@inproceedings{zhu2025patchstructdiffusion,
  title={PatchStructDiffusion: Structure-Aware Conditional Diffusion for 3D Brain MRI Inpainting},
  author={Zhu, Jiufu and Wang, Jianhang and Fan, Wenlong and Wei, Boyang and Ye, Chuhang and Li, Jiarui and Luo, Huiyu and Zhang, Lintao},
  booktitle={International Conference on Virtual Reality and Visualization (ICVRV)},
  pages={228--233},
  year={2025},
  publisher={IEEE},
  address={Bogota, Colombia},
  doi={10.1109/ICVRV67992.2025.00047}
}

@inproceedings{tao2026trustworthy,
  title={Trustworthy Longitudinal Brain MRI Completion: A Deformation-Based Approach with KAN-Enhanced Diffusion Model},
  author={Tao, Tianli and Wang, Ziyang and Yang, Delong and Zhang, Han and Zhang, Le},
  booktitle={International Symposium on Biomedical Imaging (ISBI)},
  year={2026},
  publisher={IEEE},
  address={London, United Kingdom},
  doi={10.1109/ISBI61048.2026.11515615}
}

@inproceedings{yu2025endoscopic,
  title={Endoscopic Artifact Inpainting for Improved Endoscopic Image Segmentation},
  author={Yu, Zhangyuan and Du, Chenlin and Liang, Hongrui and Zheng, Xiuqi and Ma, Zeyao and Wu, Mingjun and Ao, Mingwu and Lao, Qicheng},
  booktitle={International Conference on Medical Image Computing and Computer-Assisted Intervention (MICCAI)},
  pages={191--201},
  year={2025},
  publisher={Springer},
  address={Cham},
  doi={10.1007/978-3-032-05127-1\_19}
}

@inproceedings{you2025fb,
  title={FB-Diff: Fourier Basis-guided Diffusion for Temporal Interpolation of 4D Medical Imaging},
  author={You, Xin and Yang, Runze and Zhang, Chuyan and Jiang, Zhongliang and Yang, Jie and Navab, Nassir},
  booktitle={International Conference on Computer Vision (ICCV)},
  pages={28010--28020},
  year={2025},
  publisher={IEEE/CVF},
  address={Honolulu, HI, USA},
  doi={10.1109/ICCV51701.2025.02600}
}

@INPROCEEDINGS{Peebles2023,
  author={Peebles, William and Xie, Saining},
  booktitle={2023 IEEE/CVF International Conference on Computer Vision (ICCV)}, 
  title={Scalable Diffusion Models with Transformers}, 
  year={2023},
  volume={},
  number={},
  pages={4172-4182},
  doi={10.1109/ICCV51070.2023.00387}}

@inproceedings{morao2025data,
  title={Data Augmentation for Medical Imaging: Counterfactual Simulation of Acquisition Parameters via Conditional Diffusion Model},
  author={Mor{\~a}o, Pedro A and Forghani, Yasna and Lou{\c{c}}{\~a}o, Nuno and Gouveia, Pedro and Figueiredo, Mario AT and Santinha, Joao},
  booktitle = {International Conference on Medical Imaging with Deep Learning (MIDL)},
  pages={1164-1180},
  year={2025},
  volume = {301},
  publisher = {Proceedings of Machine Learning Research},
  address={Salt Lake City, USA},
  url={https://proceedings.mlr.press/v301/morao26a.html}
}

\end{document}